\documentclass[10pt,journal,compsoc]{IEEEtran}

\usepackage[nocompress]{cite}

\usepackage{url}
\usepackage{ragged2e}
\usepackage{epsfig}
\usepackage{graphicx}
\usepackage{amsmath,amssymb} 
\usepackage{mathptmx}      
\usepackage{algorithm}
\usepackage{algorithmicx}
\usepackage{graphics}
\usepackage{mathrsfs}
\usepackage{color}
\usepackage[normalem]{ulem}
\usepackage{multirow}
\usepackage{float}
\usepackage{amsfonts}
\usepackage{bm}
\usepackage[table]{xcolor}
\usepackage{colortbl}
\usepackage{diagbox}
\usepackage{rotating}
\usepackage{contour}
\usepackage{enumitem}
\usepackage{stfloats}
\usepackage{threeparttable}
\usepackage{times}
\usepackage{caption}
\usepackage{epsfig}
\usepackage{graphicx}
\usepackage{utfsym}
\usepackage{graphics}
\usepackage{array}
\usepackage{colortbl}
\usepackage{rotating}
\usepackage{booktabs}
\usepackage{overpic}
\usepackage{bbding}
\usepackage{makecell}
\usepackage{adjustbox}
\usepackage{textcomp}
\usepackage{subfig}
\usepackage[colorlinks=true,linkcolor=blue,citecolor=blue,breaklinks=true,urlcolor=magenta]{hyperref}

\DeclareMathAlphabet{\CMcal}{OMS}{cmsy}{m}{n}

\def\ie{\emph{i.e.}}
\def\eg{\emph{e.g.}}

\def\etal{{\em et al.~}}

\definecolor{mygray1}{gray}{.75}

\def\ie{\emph{i.e.}}
\def\eg{\emph{e.g.}}

\def\etal{\emph{et al. }}

\newcommand{\figref}[1]{Fig.~\ref{#1}}
\newcommand{\tabref}[1]{Table~\ref{#1}}
\newcommand{\equref}[1]{Eqn.~\ref{#1}}
\newcommand{\secref}[1]{\S\ref{#1}}

\definecolor{mygray4}{gray}{0.55}
\definecolor{myGreen1}{RGB}{236, 246, 239}

\newcommand{\red}[1]{{\color[HTML]{FD6864}\textbf{{#1}}}}
\newcommand{\blue}[1]{{\color[HTML]{34CDF9}\textbf{{#1}}}}

\newcommand{\Rmark}{\usym{1F5F8}}

\RequirePackage{silence}
\usepackage{amsmath,amsfonts,bm}
\usepackage{mathtools}

\def\eqref#1{equation~\ref{#1}}

\def\1{\bm{1}}

\def\va{{\mathbf{a}}}

\def\vm{{\mathbf{m}}}
\def\vn{{\mathbf{n}}}

\def\vp{{\mathbf{p}}}
\def\vq{{\mathbf{q}}}

\def\mA{{\mathbf{A}}}

\def\mD{{\mathbf{D}}}

\def\mF{{\mathbf{F}}}

\def\mI{{\mathbf{I}}}

\def\mK{{\mathbf{K}}}

\def\mR{{\mathbf{R}}}

\def\mX{{\mathbf{X}}}
\def\mY{{\mathbf{Y}}}

\DeclareMathAlphabet{\mathsfit}{\encodingdefault}{\sfdefault}{m}{sl}
\SetMathAlphabet{\mathsfit}{bold}{\encodingdefault}{\sfdefault}{bx}{n}

\def\sA{{\mathbb{A}}}

\def\sX{{\mathbb{X}}}
\def\sY{{\mathbb{Y}}}

\begin{document}

\title{Context-measure: Contextualizing Metric \\ for Camouflage}

\author{
Chen-Yang Wang,~
Ge-Peng Ji,~
Song Shao,
Ming-Ming Cheng,~
Deng-Ping Fan\\
\IEEEcompsocitemizethanks{
\IEEEcompsocthanksitem Corresponding author: Deng-Ping Fan (E-mail: \href{dengpfan@gmail.com}{dengpfan@gmail.com}).
\IEEEcompsocthanksitem Deng-Ping Fan is with NKIARI (SHENZHEN FUTIAN) \& SLAI, Shenzhen, 518045, China, and he is also with VCIP, CS, Nankai University, Tianjin, 300071, China. Chen-Yang Wang, Ge-Peng Ji, Song Shao and Ming-Ming Cheng are with VCIP, CS, Nankai University, Tianjin, 300071, China. 
Song Shao is also with Chongqing Changan Wangjiang Industrial Group Co., Ltd., Chongqing, 401120, China.
}
}

\IEEEtitleabstractindextext{%
\begin{abstract} \justifying
Camouflage relies heavily on context, but current metrics used in camouflaged object segmentation ignore contextual cues. 
We identify two major drawbacks of these metrics: first, the \textbf{Dimension Flaw} — a predicted foreground map usually contains both pixel labels and probability scores, whereas ground truth provides only \textit{one-dimensional} binary labels; second, the \textbf{Range Flaw} — these metrics struggle to capture \textit{full-range} pixel dependencies.
Thus, we propose \textbf{Context-measure}, a novel context-aware evaluation paradigm built on a probabilistic pixel correlation framework. It augments the ground truth with pixel-level contextual affinity and builds a perception cycle, achieving greater consistency with human perception.
Extensive experiments using four meta-measures show that our Context-measure comprehensively outperforms all widely adopted metrics for camouflaged object segmentation.
To our knowledge, this is the first metric designed for camouflaged scenarios.
Code is available at \href{https://github.com/pursuitxi/Context-measure}{https://github.com/pursuitxi/Context-measure}.
\end{abstract}

}

\maketitle

\IEEEdisplaynontitleabstractindextext

\IEEEpeerreviewmaketitle

\IEEEraisesectionheading{\section{Introduction}\label{sec:introduction}}

\IEEEPARstart{E}{valuation} metrics are fundamental to the image segmentation community, serving both as a basis for model benchmarking, \eg, MS COCO~\cite{lin2014microsoft}, PASCAL VOC~\cite{everingham2010pascal}, and as a catalyst for methodological innovation~\cite{zhao2019optimizing,milletari2016v,sudre2017generalised}. As a field evolves, it naturally calls for specialized evaluation metrics that align with its distinctive problem settings. This study focuses on camouflaged object segmentation (COS)~\cite{fan2020camouflaged,fan2021concealed}, where models are required to segment objects that blend into their surroundings, exhibiting low target--context contrast and high structural homogeneity. 

Despite the inherently context-dependent nature of COS, existing evaluation metrics used in COS assess model performance solely by comparing the predicted foreground map (FM) with the manually annotated ground truth (GT), while disregarding the surrounding visual context~\cite{margolin2014evaluate,fan2017structure,fan2018enhanced}. Such evaluation is therefore inherently \textit{context-blind}, raising a fundamental question: \textbf{\textit{``Can context be ignored when evaluating model performance in camouflaged scenarios?''}}

\begin{figure}[t!]
    \centering
    \includegraphics[width=1\linewidth]{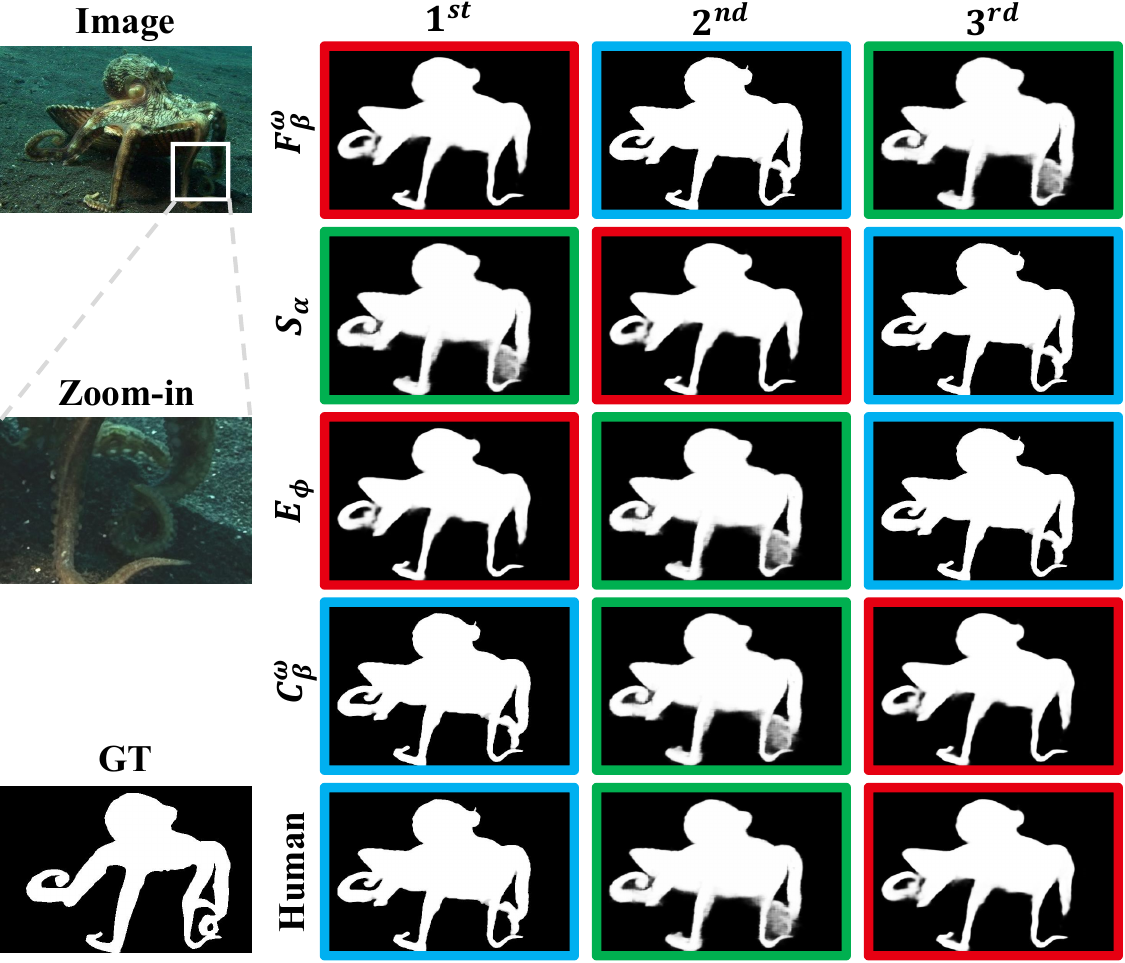}
    \caption{
    \textbf{Inaccuracy of existing COS metrics.} We examine how different evaluation metrics rank three predicted foreground maps produced by three COS models and marked with blue, green, and red borders, respectively. As indicated by the human judgment in the last row, the blue-bordered map ranks first, followed by the green- and red-bordered maps. Nevertheless, all of the widely used metrics, including $F_{\beta}^{\omega}$~\cite{margolin2014evaluate}, $S_{\alpha}$~\cite{fan2017structure}, and $E_{\phi}$~\cite{fan2018enhanced}, fail to rank these maps correctly. By contrast, the ranking produced by our $C_{\beta}^{\omega}$ is consistent with human judgment.
    }
    \label{fig:teaser_figure}
    \vspace{-13pt}
\end{figure}

To answer this question, please refer to the example shown in \figref{fig:teaser_figure}. 
The image (first column) depicts a camouflaged octopus whose appearance blends into the surrounding underwater environment. 
We use three COS models, CamoDiffusion~\cite{sun2025conditional}, VSCode-V2~\cite{luo2025vscode}, and LSR~\cite{lv2021simultaneously}, to segment the camouflaged object in the image, obtaining the blue-, green-, and red-bordered predicted maps, respectively. 
The most notable discrepancies among these three maps occur in the lower-right region.
This region contains a slender tentacle, a curled endpoint, and several narrow background gaps enclosed or separated by the tentacle. Moreover, it is partially obscured by shadows, making it particularly difficult to segment. 
In fact, the blue-bordered map provides a relatively accurate prediction of this region, whereas the green-bordered map produces only a blurry prediction, and the lower-right tentacle is barely detected in the red-bordered map. However, due to the lack of context awareness, all existing evaluation metrics fail to rank these maps correctly. This example provides a clear answer to the above question: \textbf{\textit{Context plays an indispensable role in evaluating model performance in camouflaged scenarios, which cannot be ignored.}}

\begin{figure}
    \centering
    \includegraphics[width=1\linewidth]{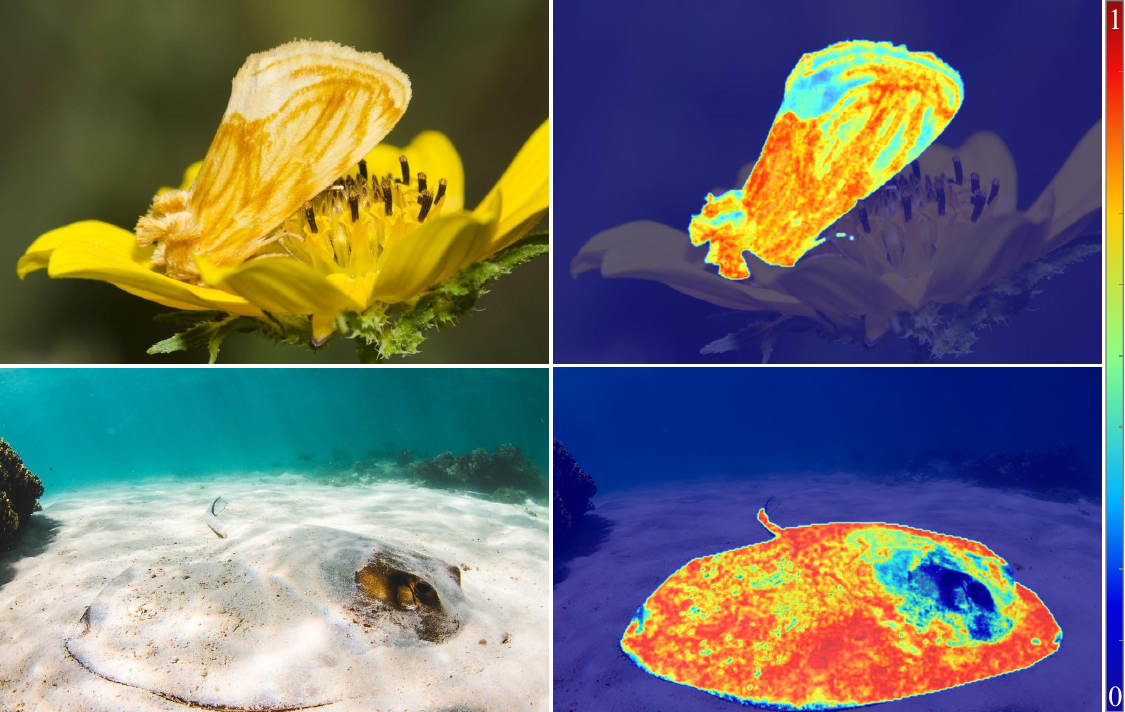}
    \caption{\textbf{Spatial variation in contextual affinity.} We visualize the pixel-level affinity map on two camouflaged objects from COD10K~\cite{fan2020camouflaged}. Warmer colors denote regions that are more similar to their surroundings and thus exhibit higher contextual affinity, whereas cooler colors denote more distinguishable regions. 
    }
    \label{fig:visualization}
\end{figure}

We identify two major drawbacks of existing metrics that result in context-blind evaluation.
The first is the \textit{Dimension Flaw}. 
Currently, almost all COS models output continuous-valued, non-binary maps (pixel values in $[0,1]$), where each pixel jointly conveys its predicted target/non-target label and the model's confidence in that prediction.
In contrast, GT masks are typically binary (pixel values in $\{0,1\}$) and provide only the target/non-target labels, without indicating how difficult that each pixel is to distinguish from its surrounding context.
Existing metrics therefore lack a GT-side basis for differentiating pixels according to their contextual difficulty. As a result, visually salient pixels and highly camouflaged pixels are treated uniformly, despite their substantially different affinities with the surrounding context. This makes existing metrics insensitive to the target--context affinity that critically determines segmentation difficulty in COS.

The absence of this second dimension is rooted in a more fundamental drawback, which we term the \textit{Range Flaw}. To characterize pixel-wise contextual difficulty, an ideal metric should capture dependencies between arbitrary pixel pairs, namely, \textit{full-range} pixel dependencies. Existing metrics fall short in two distinct ways. Some metrics (\eg, IoU~\cite{everingham2010pascal}, $F_{\beta}$~\cite{alpert2011image,arbelaez2010contour}, and $E_{\phi}$~\cite{fan2018enhanced}) do not explicitly capture inter-pixel dependencies and thus operate with \textit{zero-range} dependencies. Others (\eg, $F_{\beta}^{\omega}$~\cite{margolin2014evaluate} and $S_{\alpha}$~\cite{fan2017structure}) incorporate spatial or structural interactions only through predefined pixel relations or restricted regions, remaining limited to \textit{half-range} dependencies. Neither category can fully characterize how each pixel relates to its complete surrounding context. Consequently, the GT remains a context-blind binary reference, while pixels in the FM are evaluated without fully accounting for the rich spatial dependencies that govern camouflage.

To tackle these issues, we propose \textbf{Context-measure} ($C_{\beta}^{\omega}$), a novel context-aware evaluation paradigm. 
To address the \textit{Dimension Flaw}, we augment the binary GT mask with a pixel-wise contextual affinity that quantifies the extent to which each GT pixel visually blends with its surrounding context. As shown in \figref{fig:visualization}, the resulting affinity map provides a continuous, pixel-level estimate of contextual difficulty, supplementing the binary target label with a second dimension of information and thereby bringing the GT into correspondence with the non-binary FM.
To address the \textit{Range Flaw}, we establish a probabilistic pixel correlation framework capable of explicitly capturing dependencies between arbitrary pixel pairs. 
Built upon this framework, a \textit{perception cycle} evaluates the predicted FM from two complementary directions. 
The first proceeds from the FM to the GT, referred to as \textit{forward inference}, and estimates, for each predicted pixel, its correlation with the entire GT object, thereby measuring how much GT-relevant information is conveyed by the predicted FM. Conversely, the GT-to-FM direction, referred to as \textit{reverse deduction}, estimates, for each GT pixel, the extent to which it is captured by the predicted FM. Together, these two directions form a closed perceptual loop that evaluates segmentation quality by jointly considering the FM-to-GT and GT-to-FM relationships.

To evaluate the alignment of rankings produced by evaluation metrics with human perception, we curate \textbf{CamoHR}, a new dataset comprising 750 predicted FMs annotated with human-perceived quality rankings in camouflaged scenarios.
Compared with existing widely adopted metrics, our measure achieves a 41\% relative improvement in consistency with human judgment. 
Furthermore, extensive experiments across four meta-measures~\cite{margolin2014evaluate} show that Context-measure comprehensively outperforms all widely adopted evaluation metrics. To our knowledge, this is the first metric specifically tailored for camouflaged scenarios.

In summary, our contributions are three-fold:
\begin{itemize}[leftmargin=1.5em]
    \item We identify two major drawbacks of existing context-blind metrics -- the Dimension Flaw and the Range Flaw -- and analyze how they collectively lead to context-blind evaluation in camouflaged scenarios.
    \item We propose a pixel-wise contextual affinity to resolve the Dimension Flaw, and establish a FM-GT-FM perception cycle built upon a probabilistic pixel correlation framework to address the Range Flaw, achieving context-aware evaluation.
    \item We develop \textbf{Context-measure}, the first COS-specific evaluation metric, and construct the CamoHR dataset to benchmark metric consistency with human perception, extensively validating its effectiveness across multiple experiments.
\end{itemize}

\section{Related Studies}\label{sec:related_studies}

\subsection{Revisiting  Metrics}\label{sec:revisiting_metrics}

    We categorize existing evaluation metrics into three groups according to the granularity of visual information they assess.
    
    \noindent\textbf{Pixel-aware Metrics.} 
    These metrics derive their scores from pixel-wise FM--GT comparisons or aggregated pixel-level statistics, without explicitly capturing dependencies among pixels.
    Mean absolute error ($\CMcal{M}$)~\cite{perazzi2012saliency} computes the average absolute difference between the FM and GT over all pixels. Intersection over Union (IoU)~\cite{everingham2010pascal} measures the set overlap between the FM and GT object regions by aggregating their pixel-level intersection and union. The F-measure ($F_{\beta}$)~\cite{alpert2011image,arbelaez2010contour}, defined as the harmonic mean of precision and recall, balances these two aspects to evaluate overall performance. When $\beta=1$, it reduces to the F1 score, which is mathematically related to the Jaccard Index (JI) \cite{jaccard1901etude}. Notably, they satisfy the relation of $\text{IoU}=\text{JI}=\frac{\text{F1}}{2-\text{F1}}$, thus yielding identical rankings and often being used interchangeably. The multiscale IoU (mIoU) \cite{ahmadzadeh2021multiscale} extends IoU by incorporating multiple resolutions, enabling more comprehensive evaluation across scales and improving sensitivity to fine boundary details.

    \noindent\textbf{Region-aware Metrics.}
    Unlike pixel-aware metrics, region-aware metrics incorporate spatial relations within predefined regions or between specific pixel pairs. Margolin \etal\cite{margolin2014evaluate} identify two perceptual limitations of the $F_{\beta}$, termed the E-Flaw and D-Flaw, and propose the weighted F-measure ($F_{\beta}^{\omega}$) to alleviate them through distance-based error weighting and dependency modeling among prediction errors. The Structure-measure ($S_{\alpha}$)~\cite{fan2021structure,fan2017structure} integrates both region- and object-aware components, providing a more holistic evaluation of structural integrity in the FM. The Size-invariance MAE ($\CMcal{M}_{SI}$) \cite{li2024sizeinvariance} measures prediction errors in an object-wise manner so that objects of different sizes contribute equally to the final score.

    \noindent\textbf{Image-aware Metrics.} 
    These metrics incorporate image-level statistics to evaluate the overall alignment between FM and GT.
    The enhanced-alignment measure ($E_{\phi}$)~\cite{fan2018enhanced,fan2021cognitive} combines local pixel-level alignment with global image-level statistics, yielding results that are more consistent with human visual perception of foreground segmentation quality.

    A summary of the key features of the evaluation metrics discussed above can be found in \tabref{tab:summary-relatedwork}.
    
\begin{table}[b!]
    \centering
    \renewcommand{\arraystretch}{1}
    \renewcommand{\tabcolsep}{0.8mm}
    \caption{
    \textbf{Comparison of evaluation metric properties.} 
    \underline{\smash{Type}} indicates whether a metric evaluates only binary masks (Bin.) or can also evaluate continuous predicted maps (Non-bin.).
    \underline{\smash{Asm}} indicates whether the FM and GT play asymmetric roles in the evaluation, such that exchanging them may change the resulting score.
    \underline{\smash{E-Flaw}} (equal-importance flaw~\cite{margolin2014evaluate}) refers to the tendency of a metric to penalize all erroneous pixels equally, regardless of perceptual significance. \underline{\smash{D-Flaw}} (dependence flaw~\cite{margolin2014evaluate}) refers to the assumption of pixel-wise independence, which causes spatial correlations to be ignored. \underline{\smash{Image/Region/Pixel}} specifies the perceptual hierarchy where a metric operates. 
    \underline{\smash{Cam}} indicates whether a metric was originally designed for camouflaged scenarios.
    }
    \label{tab:summary-relatedwork}
    \begin{tabular}{rcccccccc}
    \toprule
    Metric  & Type & Asm & E-Flaw & D-Flaw & Image & Region & Pixel & Cam  \\ 
    \midrule
    $\CMcal{M}$~\cite{perazzi2012saliency}    &   Non-bin.    &         &                &                &             &                &  \Rmark             &  \\
    IoU/JI~\cite{jaccard1901etude}     &   Bin.       &                &                &                &             &                &    \Rmark        &  \\
    $F_{\beta}$~\cite{alpert2011image,arbelaez2010contour}      &   Bin. &                &                &                &           &                &   \Rmark               &  \\
    mIoU~\cite{ahmadzadeh2021multiscale}    &   Bin.     &                &                &                &             &                &    \Rmark        & \\
     $F_{\beta}^{\omega}$~\cite{margolin2014evaluate} &   Non-bin. &  \Rmark & \Rmark &  \Rmark &  &  \Rmark & \Rmark  & \\
    
    $S_{\alpha}$~\cite{fan2017structure,fan2021structure}  &   Non-bin.   & \Rmark             &~~\Rmark$^\star$            &~~\Rmark$^\star$              &   \Rmark             &\Rmark              &          & \\
    $\CMcal{M}_{SI}$~\cite{li2024sizeinvariance}  &   Non-bin.   & \Rmark   &     &              &             &\Rmark              &\Rmark          & \\
    $E_{\phi}$~\cite{fan2018enhanced,fan2021cognitive}  &   Bin.   &                &                &                & \Rmark             &                &\Rmark         &\\ 
    \midrule
    \rowcolor[HTML]{EFEFEF}
    $C_{\beta}^{\omega}$ (Ours) &   Non-bin. &  \textbf{\Rmark}             &\textbf{\Rmark}              &\textbf{\Rmark}              & \textbf{\Rmark}             &\textbf{\Rmark}              &\textbf{\Rmark}      &\textbf{\Rmark}     \\ 
    \bottomrule
    \end{tabular}
    \begin{tablenotes}
    \item $\star$ indicates that $S_\alpha$ addresses E-Flaw \& D-Flaw when erroneous pixels are distributed across regions; otherwise, it fails to handle either.
    \end{tablenotes}
\end{table}

\begin{figure*}[t]
    \centering
    \includegraphics[width=1\linewidth]{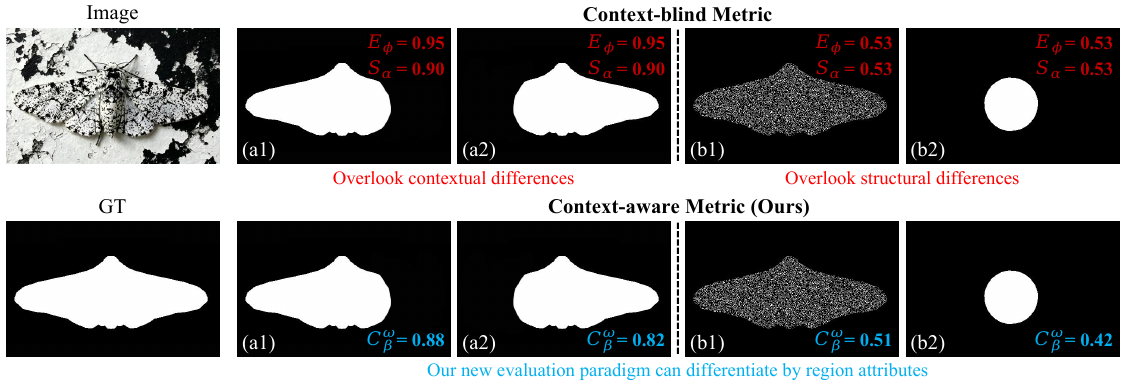}
    \vspace{-18pt}
    \caption{\textbf{Flaws of context-blind metrics in camouflage.} Compared with the two most widely used metrics ($E_{\phi}$ \& $S_{\alpha}$), our Context-measure ($C_{\beta}^{\omega}$) more effectively distinguishes masks in camouflaged scenarios, aligning more closely with human perception. 
    }
    \label{fig:metric_vulnerability}
\end{figure*}

\subsection{Metric Flaws in COS}\label{sec:fairness_of_saliency_metric}

    Although the aforementioned metrics have their own characteristics, they share several fundamental flaws when applied to COS.

    \noindent\textbf{Dimension Flaw.}
    GT masks typically provide only binary labels, without indicating the contextual difficulty of each pixel.
    Consequently, existing metrics cannot differentiate geometrically equivalent FMs according to the contextual difficulty of the regions they segment. This limitation is illustrated in \figref{fig:metric_vulnerability}. The predicted maps in (a1) and (a2) exhibit geometrically symmetric patterns relative to the GT and hence receive identical scores from all current evaluation metrics. However, the region correctly segmented in (a1) exhibits weaker contrast with the surroundings, whereas its counterpart in (a2) is more visually distinctive and easier to segment. 
    These two maps therefore involve different levels of contextual difficulty and, cognitively, should not be regarded as equivalent in quality or assigned identical scores.
    By augmenting the GT with pixel-wise contextual affinity, our $C_{\beta}^{\omega}$ captures this contextual distinction and assigns different scores to them.

    \noindent\textbf{Range Flaw.} 
    From a dependency-based perspective, metrics can capture \textit{zero-range}, \textit{half-range}, or \textit{full-range} dependencies. 
    Zero-range metrics treat pixels independently, half-range metrics capture only predefined or spatially restricted dependencies, whereas full-range metrics consider dependencies between arbitrary pixel pairs.
    Metrics such as $\CMcal{M}$, IoU/JI, $F_\beta$, and $E_{\phi}$ treat all pixels as mutually independent units, and are therefore categorized as zero-range metrics. Other metrics partially incorporate spatial information but remain limited to half-range dependencies. For example, $F_{\beta}^{\omega}$ captures dependencies only among error pixels, whereas $S_{\alpha}$ captures dependencies only among pixels within the same subregion, with the four subregions divided by the horizontal and vertical axes passing through the target centroid. Such restricted relationships remain insufficient to characterize how each pixel interacts with its complete surrounding context.
    This limitation is further illustrated by the predicted maps in \figref{fig:metric_vulnerability} (b1) and \figref{fig:metric_vulnerability} (b2). These two maps are constructed to preserve the regional statistics captured by existing metrics, including the number of target pixels within each centroid-defined quadrant, while exhibiting distinctly different spatial arrangements. 
    Hence both $E_{\phi}$ and $S_{\alpha}$ assign them identical scores. In contrast, by capturing full-range dependencies, our $C_{\beta}^{\omega}$ can distinguish their spatial relationship and assigns different scores to them.
    
\section{Proposed Context-measure}\label{sec:cmeasure}

In this section, we first augment the GT with a contextual affinity map (\secref{sec:affinity}). Then we establish a probabilistic pixel correlation framework (\secref{sec:definition}) that explicitly models the dependency between arbitrary pixel pairs, thereby providing the necessary foundation for context-aware evaluation. Finally, building on this foundation, we construct a perception cycle (\secref{sec:loop}) that evaluates segmentation quality from two complementary directions: the \textit{forward inference} and the \textit{reverse deduction}.

Throughout the paper, we denote vectors as $\va$, matrices as $\mA$, sets as $\sA$, and equality by definition as $\coloneqq$. Let $\mX$ and $\mY$ denote the predicted FM and GT mask, respectively, and let $\mI$ denote the original image. For each pixel $\vp$, $\mX(\vp)\in[0,1]$ and $\mY(\vp)\in\{0,1\}$ denote its values in $\mX$ and $\mY$, respectively. We define two target pixel sets: $\sX_t\coloneqq\{\vp\mid \mX(\vp)>0\}$ and $\sY_t\coloneqq\{\vp\mid \mY(\vp)>0\}$, representing the predicted and GT target regions.

\subsection{Pixel-wise Contextual Affinity}\label{sec:affinity}

As discussed in the Introduction \secref{sec:introduction}, GT provides only binary labels, missing the contextual difficulty 
information. To resolve this Dimension Flaw, we propose to augment GT with a \textit{contextual affinity} map $\mD$, in which each value $\mD(\vp)$ quantifies how strongly the GT pixel $\mathbf{p}$ blends into its surrounding context. 

In camouflaged scenarios, contextual affinity is primarily reflected by how strongly a target region blends into its surrounding context. 
Existing camouflage quantification methods typically measure this property at the object level~\cite{lamdouar2023making,das2025camouflage}.
However, $\mD$ requires a pixel-level estimate, because camouflage is spatially non-uniform: different regions of the same target may exhibit substantially different degrees of surrounding context matching. 
While visual blending can in principle be characterized along multiple dimensions -- such as texture, depth, and learned features -- color similarity stands out as the most direct, interpretable, and independently verifiable indicator of local target--context affinity in natural scenes, and its central role in camouflage is well-established~\cite{stevens2009animal}. We therefore estimate $\mD$ based on pixel-level color similarity, measured in the CIELAB color space~\cite{luo2001development}. This is not merely a convenient choice: RGB distances are device-oriented and do not reliably reflect human color discrimination, whereas CIELAB was specifically designed to be perceptually uniform, making its distances a principled proxy for how visually similar two colors appear to a human observer~\cite{mokrzycki2011colour,luo2001development}. 
Next, we detail how $\mD$ is estimated in two stages.

    \begin{figure}[t!]
        \centering
        \includegraphics[width=1\linewidth]{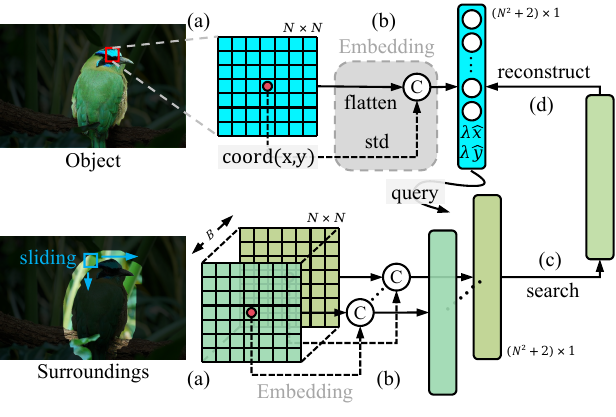}
        \caption{\textbf{Illustration of contextual reconstruction.} 
        Object regions are first dilated to define their contextual surroundings. Overlapping patches are then extracted separately from the object and contextual regions.
        Each patch is embedded as a feature vector by combining LAB values with scaled spatial coordinates. For each object vector, use \textit{ANN} to find the best contextual match, which is projected back to reconstruct the object.}
        \label{fig:reconstruct}
    \end{figure}

\noindent\textbf{Stage I: Contextual Reconstruction.} We perform this stage in four steps. 
\textit{(a) Extraction.} Unlike Lamdouar \etal \cite{lamdouar2023making}, we begin by expanding the object region outward by $W$ pixels through morphological dilation~\cite{haralick1987image}, thereby defining the resulting band as the contextual surroundings of the object. 
We then densely extract all overlapping $N \times N$ patches from both the object region and its contextual surroundings.
\textit{(b) Embedding.} Each patch is embedded as a feature vector, obtained by concatenating its flattened LAB color values with standardized spatial coordinates, the latter scaled by a factor $\lambda$ to balance spatial and chromatic contributions. 
\textit{(c) Search.} For each object feature vector, we perform an approximate nearest neighbor (\textit{ANN}) search over the $B$ contextual surrounding feature vectors to find the most compatible one, thereby obtaining a spatially coherent contextual correspondence.
\textit{(d) Reconstruction.} Each matched feature vector is projected back into image space, replacing the original object region with its contextually harmonized reconstruction. This stage is illustrated in \figref{fig:reconstruct}.

\noindent\textbf{Stage II: Affinity Mapping.} Given the reconstructed object region, we quantify the contextual affinity $\mD$ by measuring the color difference between the original and reconstructed pixels in the LAB color space. Specifically, these differences are transformed through a nonlinear mapping to obtain $\mD$:
\begin{equation}
    \mD \coloneqq \frac{\exp\Big(\gamma\cdot\Big(\mathbf{1} - 
    \frac{1}{100}\Delta(\mR_{lab}, 
    \mI_{lab})\Big)\Big) - \mathbf{1}}{\exp(\gamma) - 1},
\end{equation}
where $\gamma$ modulates the nonlinearity, and $\Delta$ denotes the CIEDE2000 color difference~\cite{luo2001development} between the 
computed reconstruction $\mR_{lab}$ and the original object 
$\mI_{lab}$ in LAB space. A smaller difference indicates stronger blending, yielding a higher affinity value. The visualized heatmaps of $\mD$ are presented in \figref{fig:visualization}.

\subsection{Probabilistic Pixel Correlation}\label{sec:definition}

To address the Range Flaw, we establish a probabilistic framework that explicitly captures full-range pixel dependencies. The key design principle is that pixel correlation should decay with spatial distance: pixels that are farther apart are less likely to be structurally related. We model this using a Gaussian formulation conditioned on the GT object $\sY_t$, which encodes the spatial structure of the target object.

Specifically, we estimate the covariance matrix $\boldsymbol{\Sigma}$ from $\sY_t$, capturing the shape of the GT object. To ensure that pixel-wise correlations remain comparable across images regardless of resolution, we normalize the covariance as $\hat{\boldsymbol{\Sigma}}=\frac{\alpha^2}{\operatorname{Tr}(\boldsymbol{\Sigma})}\boldsymbol{\Sigma}$, where $\alpha$ is a scalar hyperparameter and $\operatorname{Tr}(\cdot)$ denotes the trace operator. The correlation between any two pixels $\vm$ and $\vn$ is then defined as:
\begin{equation}\label{equ:pixel_correlation}
    P(\vm,\vn) \coloneqq \frac{\exp{\Big(-\frac{1}{2}
    (\vn-\vm)^{\top}\hat{\boldsymbol{\Sigma}}^{-1}
    (\vn-\vm)\Big)}}{2\pi\sqrt{|\hat{\boldsymbol{\Sigma}}|}},
\end{equation}
where $\hat{\boldsymbol{\Sigma}}^{-1}$ and $|\hat{\boldsymbol{\Sigma}}|$ 
denote the inverse and determinant of $\hat{\boldsymbol{\Sigma}}$. 
$P(\vm, \vn)$ quantitatively measures the strength of the relationship between the two pixels: the farther apart they are, the weaker their 
correlation becomes, following a Gaussian decay.

\subsection{FM-GT-FM Perception Cycle}\label{sec:loop}

\begin{figure*}[t!]
    \centering
    \begin{overpic}[width=1\linewidth]{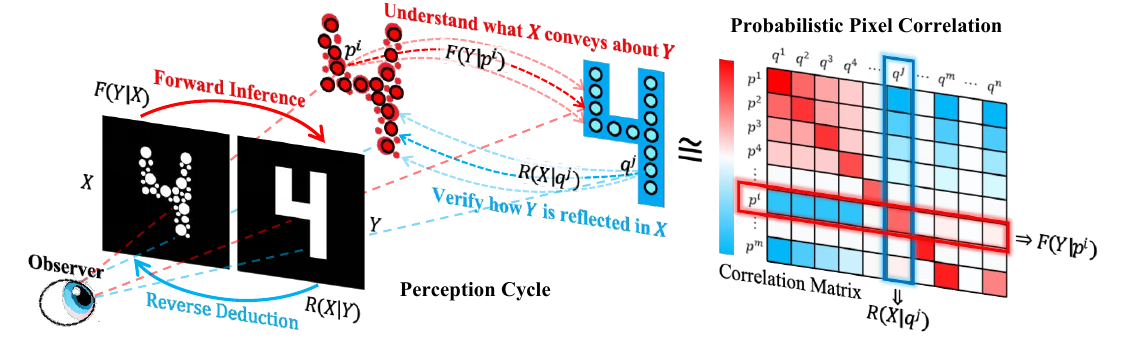}
        \put(90,28.14){(\secref{sec:definition})}
        \put(50,4.6){(\secref{sec:loop})}
    \end{overpic}
    \vspace{-21pt}
    \caption{\textbf{Our perception cycle framework.} We formulate the predicted map evaluation problem as a \textit{perception cycle}, which alternates between two processes: \textit{forward inference} $\mF(\mY|\mX)$ and \textit{reverse deduction} $\mR(\mX|\mY)$. Intuitively, the former estimates what the predicted map conveys about reality, and the latter verifies how reality is reflected in the predicted map. Details are provided in \secref{sec:loop}.
    }
    \label{fig:framework}
\end{figure*}

With the augmented ground truth and the pixel correlation $P(\cdot,\cdot)$ in place, we now construct the perception cycle to complete the evaluation. When evaluating segmentation quality, human observers typically engage in an iterative comparison -- shifting attention back and forth between FM and GT to assess their consistency. This inspires a loop alternating between two complementary directions: \textit{forward inference} $\mF(\mY|\mX)$, which begins from the predicted FM and is weighted by the predicted probabilities, and \textit{reverse deduction} $\mR(\mX|\mY)$, which begins from the GT and is weighted by the contextual affinity $\mD$. The framework is illustrated in \figref{fig:framework}. We next detail the implementation of each direction.

\noindent\textbf{Forward Inference.}
Specifically, for each predicted target pixel $\vp^i \in \sX_t$, forward inference estimates its correlation with the GT target as a whole, weighted by the model's predicted probability $\mX(\vp^i)$. This captures how much information the predicted FM conveys about the GT, with higher-confidence predictions contributing more. The corresponding formulation is given by:
\begin{equation}
    \mF(\mY|\vp^i) \coloneqq \mX(\vp^i)\cdot
    \sum_{j;\vq^j\in\sY_t} P(\vp^i,\vq^j).
\end{equation}
Considering computational cost, we discretize $P(\cdot,\cdot)$ and implement it as a convolution kernel $\mK$, following the $3\sigma$ principle to cover more than 99\% of the probability mass. Under this approximation, forward inference is efficiently rewritten as:
\begin{equation}
    \mF(\mY|\mX) \approx \mX \odot (\mK \ast 
    \mY),
\end{equation}
where $\odot$ is element-wise multiplication and $\ast$ denotes convolution. The forward term is then normalized  to obtain a comparable score:
\begin{equation}
    F_m \coloneqq \|\mF(\mY|\mX)\|_1/\|\mX\|_1.
\end{equation}

\noindent\textbf{Reverse Deduction.}
Specifically, for each GT target pixel $\vq^j \in \sY_t$, it estimates the extent to which it is recovered by the predicted FM. Whereas the predicted probability naturally serves as the weight on the FM side, GT pixels carry no inherent weights -- this is precisely the Dimension Flaw we resolved in \secref{sec:affinity}. We therefore incorporate the contextual affinity $\mD(\vq^j)$ as the weight: GT pixels that blend more strongly into the context are harder to detect and thus receive greater weights in the evaluation. Formally:
\begin{equation}
    \mR(\mX|\vq^j) \coloneqq \Big(1+\mD(\vq^j)\Big) \cdot \Big (1-\prod_{i;\vp^i\in 
    \sX_t}\Big[1-\mX(\vp^i)\cdot 
    P(\vp^i,\vq^j)\Big]\Big).
\end{equation}
Similarly, $P(\cdot,\cdot)$ is approximated using a kernel $\mK_j$ centered at $\vq^j$:
\begin{align}
    1-\prod_{i;\vp^i\in 
    \sX_t}\Big[1&-\mX(\vp^i)\cdot 
    P(\vp^i,\vq^j)\Big] \nonumber \\ 
    &=1-\exp{\Big(\sum_{i;\vp^i\in \sX_t}
    \ln{\Big[1-\mX(\vp^i)\cdot 
    P(\vp^i,\vq^j)\Big]}\Big)} \nonumber \\
    &\approx 1-\exp{\Big(-\sum_{i;\vp^i\in \sX_t}
    \mX(\vp^i)\cdot 
    P(\vp^i,\vq^j)\Big)} \label{eq:Taylor}\\
    &\approx 1-\exp{\Big(-\mK_j\ast \mX\Big)}, 
    \label{eq:Approx}
\end{align}
where the approximation in \equref{eq:Taylor} relies on the first-order Taylor expansion $\ln{(1-x)}\approx -x$, valid when $\mX(\vp^i)\cdot P(\vp^i,\vq^j)$ is close to zero, which holds across most experimental samples. Since the output of \equref{eq:Approx} lies in $[0, 1-e^{-1}]$, we apply a normalization factor $\frac{e}{e-1}$ to rescale it to $[0,1]$:
\begin{equation}\label{equ:reverse_deduction}
    \mR(\mX|\mY) \approx \frac{e}{e-1} \cdot (\mathbf{1}+\mD) \odot \mY \odot 
    \Big[\mathbf{1}-\exp{\Big(-\mK\ast \mX\Big)}\Big],
\end{equation}
where $\mathbf{1}$ is an all-ones matrix and $\exp(\cdot)$ is applied element-wise.
Accordingly, the weighted reverse term is computed as:
\begin{equation}
    R_\omega \coloneqq \|\mR(\mX|\mY)\|_1/\|\mY+\mD\|_1.
\end{equation}
The two directions are integrated via harmonic weighting to form the final Context-measure:
\begin{equation}\label{equ:camo_c_measure}
    C_{\beta}^{\omega} \coloneqq \frac{(1+\beta^2)\cdot F_m \cdot R_\omega}
    {\beta^2 \cdot F_m + R_\omega},
\end{equation}
where $\beta$ controls the the relative importance of $F_m$ and $R_\omega$.

\section{Meta-Measure Experiments}\label{sec:meta}

To evaluate the effectiveness of Context-measure in comparison to existing metrics, we employ four meta-measures~\cite{margolin2014evaluate}. 

\noindent\textbf{Experimental Setup.} All competing metrics are evaluated using their default configurations. As predicted FMs are non-binary, some metrics (IoU, $F_{\beta}$, and $E_{\phi}$) require binary inputs, adaptive thresholding~\cite{achanta2009frequency} (twice the mean predicted values) is applied to binarize the FMs before evaluation. Our $C_{\beta}^{\omega}$ was configured with $\alpha=6$, $\beta^2=1.2$, $N=7$, $W=20$, $\gamma=8$, and $\lambda=20$.

\noindent\textbf{Datasets \& Models.} Experiments were conducted on three widely used benchmarks: COD10K~\cite{fan2020camouflaged}, NC4K~\cite{lv2021simultaneously}, and Trans10K~\cite{xie2020segmenting}, selected for their comprehensive coverage of diverse object types and scene complexities. For COD10K and Trans10K, models were trained on their respective official training sets and evaluated on the corresponding test sets. For NC4K, which provides only a test set, models evaluated on NC4K were trained on the COD10K training set. Seven models were used to generate the predicted FMs: FEDER~\cite{he2023camouflaged}, FSPNet~\cite{huang2023feature}, HetNet~\cite{he2023efficient}, HitNet~\cite{hu2023high}, SAM2~\cite{ravi2024sam2}, SINet-V2~\cite{fan2021concealed}, and ZoomNet~\cite{pang2022zoom}.

\subsection{Meta-Measure 1: Human Ranking}\label{sec:mm1}

We regard human judgment as one of the most important criteria for assessing segmentation quality, and an ideal evaluation metric should remain consistent with human judgments across diverse scenarios.
To assess this property, we adopt the meta-measure (MM\#1), proposed by Fan \etal\cite{fan2017structure}, to measure the consistency between metric-induced rankings and human perception.

\begin{figure}[t!]
    \centering
    \includegraphics[width=1\linewidth]{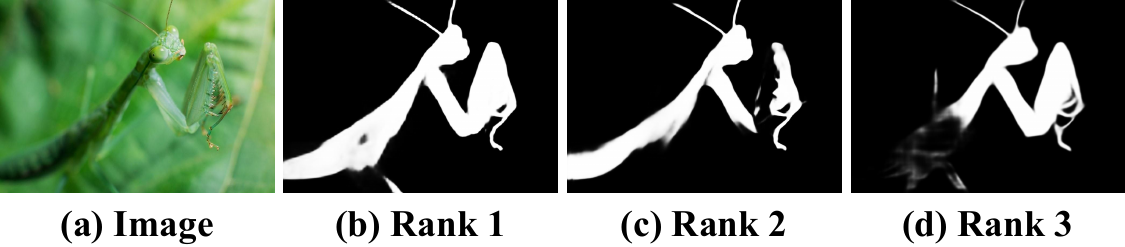}
    \caption{
    \textbf{Meta-measure 1: human ranking.} 
    Qualitative examples from our newly constructed CamoHR dataset. Each example presents a camouflaged image together with three predicted maps of varying quality. Their relative rankings within each individual example are determined by human judgments. These annotations provide perceptual supervision for evaluating the consistency between metric-induced rankings and human preferences.
    }
    \label{fig:camohr}
\end{figure}

\noindent\textbf{Ranking Data Curation.} To the best of our knowledge, there is currently no dataset in the camouflage domain that provides human-annotated rankings for model predictions. To construct such a novel human-ranked camouflage dataset, we proceed in three steps. 
\textit{(a) Image Selection.} We randomly sample images from the COD10K test set and segment them using the seven COS models introduced in the Datasets \& Models subsection. 
These models are selected to span a broad range of overall performance levels, thereby producing segmentation results of sufficiently diverse quality for reliable human ranking. 
For each sampled image, three segmentation results of varying quality are selected from the corresponding model outputs. 
\textit{(b) FM Composition.} Each predicted map is overlaid on its corresponding original image through alpha blending, with the original image and GT mask provided as references. 
\textit{(c) User Study.} We recruited 15 human participants with basic knowledge of visual perception. Each participant performed pairwise comparisons among the predicted maps and cast votes accordingly. The collective voting results were validated through discussion to reach a final consensus ranking. From a candidate pool of over 5,000 raw samples, we curated the \textbf{CamoHR} dataset, comprising 750 high-quality predicted maps annotated with human-perceived quality rankings, structured around comparative triplets where each group contains three distinct predicted foreground maps with their corresponding original image and GT mask. Then the 750 samples were divided into a validation set and a test set in a 1:4 ratio, where the validation set is used exclusively for hyperparameter selection. Examples from CamoHR are shown in \figref{fig:camohr}.

\noindent\textbf{Evaluation Protocol.} We compute Spearman's rank correlation coefficient $\rho$~\cite{best1975algorithm} to quantify the consistency. For easier interpretation, we follow \cite{margolin2014evaluate} in defining $\theta = 1 - \rho$, where a lower value indicates greater ranking consistency.

\begin{figure}[t!]
    \centering
    \includegraphics[width=1\linewidth]{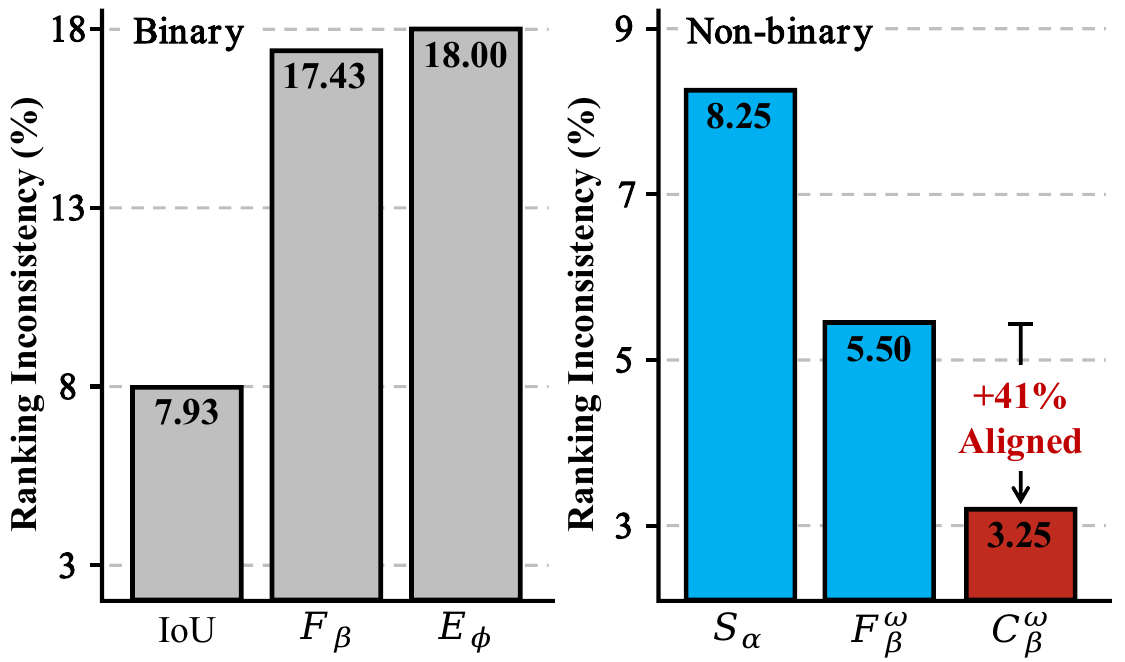}
    \vspace{-18pt}
    \caption{\textbf{Meta-measure 1: results.} We evaluate the agreement between metric-induced rankings and human judgments using $1-\rho$, where $\rho$ denotes Spearman's rank correlation coefficient~\cite{best1975algorithm}. 
    A lower value indicates stronger consistency with human perception. 
    Our $C_{\beta}^{\omega}$ achieves the lowest score, showing superior alignment with human-annotated quality rankings.}
    \label{fig:mm1}
\end{figure}

\noindent\textbf{Result Analysis.} As shown in \figref{fig:mm1}, existing metrics exhibit varying degrees of inconsistency with human judgment. Consider $E_{\phi}$ as an example. Although it integrates both global statistics and local pixel matching, its global modeling is based merely on mean energy and lacks spatial positional information. In camouflaged scenes, predicted FMs can often be globally consistent but locally mismatched, allowing background regions to receive undesirably high scores and leading to evaluation outcomes misaligned with human perception. In contrast, our $C_{\beta}^{\omega}$ achieves the strongest consistency with human judgment on CamoHR, demonstrating a 41\% relative improvement over the best-performing existing metric $F_{\beta}^{\omega}$. This validates that incorporating contextual affinity into the evaluation leads to assessments that more faithfully reflect human perception of camouflage difficulty.

\begin{figure}[b!]
    \centering
    \includegraphics[width=1\linewidth]{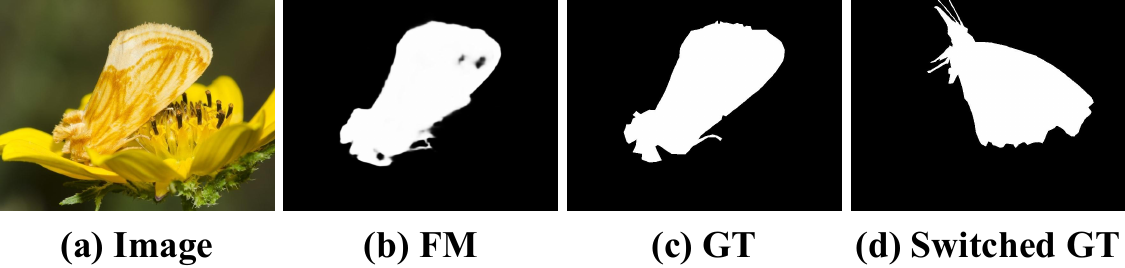}
    \caption{\textbf{Meta-measure 2: ground truth switch.} Using an incorrect paired GT as the reference should lead to a lower score for the FM segmented from (a). However, $F_{\beta}$, $F_{\beta}^{\omega}$, $S_{\alpha}$, and $E_{\phi}$ each exhibit a notably high error rate on at least one of the three datasets, occasionally assigning a higher score to the mismatched FM--GT pair. Our measure achieves the lowest error rates across all datasets and correctly yields a lower score under the incorrect reference.}
    \label{fig:mm2}
\end{figure}

\subsection{Meta-Measure 2: Ground Truth Switch}\label{sec:mm2}

In real-world applications, mismatches between predicted FMs and non-corresponding GT masks may occur due to oversights in data processing or evaluation procedures, as shown in \figref{fig:mm2}. A reliable evaluation metric should be sensitive enough to reflect such semantic mismatches. In other words, even if a predicted foreground map is of high visual quality, its score should drop significantly if the associated GT mask is semantically incorrect. 
To evaluate this property, we adopt the meta-measure (MM\#2) proposed by Margolin \etal\cite{margolin2014evaluate}, which compares the sensitivity of different metrics to entirely mismatched GT masks.

\noindent\textbf{Evaluation Protocol.} We select high-quality predicted FMs (F1 score $\ge$ 0.6) from all experimental samples, then proceed as follows. \textit{(a) Pseudo-GT Generation.} A fully permuted sequence of GT masks is constructed such that none corresponds to its original predicted FM, achieved via iterative random reassignment. \textit{(b) Dimension Alignment.} All pseudo-GTs are resized to match the dimensions of the predicted FM. \textit{(c) Error Recording.} Each predicted FM is evaluated against both its correct GT and the pseudo-GT. The error rate is defined as the proportion of cases where the pseudo-GT yields a higher score than the correct GT.

\noindent\textbf{Result Analysis.} As shown in \tabref{tab:result-mm}, columns 3,4,5, most metrics exhibit good sensitivity to GT semantic mismatches. However, $E_{\phi}$ remains susceptible to high-similarity background regions, which compromises its ability to detect semantic-level discrepancies. Our $C_{\beta}^{\omega}$ achieves near-zero error rates across all three datasets, presenting robust sensitivity to semantic mismatches.

\begin{table*}[t!]
    \centering
    \caption{
    \textbf{Quantitative analysis of metrics across four meta-measures.} 
    Unless otherwise specified, the best results are highlighted in \textbf{bold}, and values no greater than 0.01\% are reported as $^{*}$0.01\%. MM: Meta-Measure.
    }
    \label{tab:result-mm}
    \setlength\tabcolsep{3.7pt}
    \renewcommand\arraystretch{0.9}
    \vspace{-3pt}
    \begin{tabular}{cccccccccccccc}
    \toprule
    \multirow{2}{*}{Metric} & \multicolumn{1}{c}{MM\#1} & \multicolumn{3}{c}{MM\#2} & \multicolumn{3}{c}{MM\#3} & \multicolumn{3}{c}{MM\#4: Erode} & \multicolumn{3}{c}{MM\#4: Dilate}\\ \cmidrule(lr){2-2} \cmidrule(lr){3-5} \cmidrule(lr){6-8} \cmidrule(lr){9-11} \cmidrule(lr){12-14}
     & CamoHR & \multicolumn{1}{c}{COD10K} & \multicolumn{1}{c}{NC4K} & \multicolumn{1}{c}{Trans10K} & \multicolumn{1}{c}{COD10K} & \multicolumn{1}{c}{NC4K} & \multicolumn{1}{c}{Trans10K} & \multicolumn{1}{c}{COD10K} & \multicolumn{1}{c}{NC4K} & \multicolumn{1}{c}{Trans10K} & \multicolumn{1}{c}{COD10K} & \multicolumn{1}{c}{NC4K} & \multicolumn{1}{c}{Trans10K} \\ 
    \midrule 
    IoU/F1/JI  & 7.93\%  & 0.05\%  & 0.06\%  & 0.08\%  & 0.55\%  & 0.39\%  & 3.00\%  & 2.76\%  & 1.99\% & 0.48\%  & 1.67\%  & 1.32\%   & 0.46\% \\
    $F_\beta$  & 17.43\%~~ & 0.34\%  & 0.32\%  & 0.14\%  & 1.48\%  & 2.00\%  & 7.65\%   & 3.28\%   & 2.54\%  & 0.46\%  & 2.31\%   & 1.46\%  & 0.40\%  \\
    $F_{\beta}^{\omega}$  & 5.50\%  & 0.09\%                         & 0.13\%                         & 0.05\%                         & \multicolumn{1}{l}{\textbf{$^{*}$0.01\%~~}} & \multicolumn{1}{l}{\textbf{$^{*}$0.01\%~~}} & \multicolumn{1}{l}{~\textbf{$^{*}$0.01\%}} & 2.12\%                         & 1.49\%                         & 0.44\%                         & 1.13\%                    & 0.83\%                    & 0.24\%                    \\
    $S_\alpha$                                & 8.25\%     & 0.09\%                         & 0.06\%                         & 0.54\%                         & 10.47\%~~                        & 8.43\%                         & 0.53\%                        & 1.34\%                         & 0.97\%                         & 0.31\%                         & 0.85\%                    & 0.77\%                    & 0.20\%                    \\
    $E_\phi$                                  & 18.00\%~~     & 3.46\%                         & 2.37\%                         & 1.62\%                         & 13.73\%~~                        & 15.64\%~~                        & 5.62\%                         & 1.79\%                         & 1.16\%                         & 0.38\%                         & 1.16\%                    & 0.93\%                    & 0.18\%                    \\ \midrule 
    \rowcolor[HTML]{EFEFEF}
    $C_{\beta}^{\omega}\ (\text{Ours})$ & \textbf{3.25\%} & \textbf{0.02\%} & \textbf{0.03\%} & \textbf{0.03\%} & \multicolumn{1}{l}{\textbf{$^{*}$0.01\%~~}} & \multicolumn{1}{l}{\textbf{$^{*}$0.01\%~~}} & \multicolumn{1}{l}{\textbf{~$^{*}$0.01\%}} & \textbf{1.21\%} & \textbf{0.80\%} & \textbf{0.29\%} & \textbf{0.80\%} & \textbf{0.64\%} & \textbf{0.14\%}\\ \bottomrule 
    \end{tabular}
\end{table*}

\begin{figure}[t!]
    \centering
    \includegraphics[width=1\linewidth]{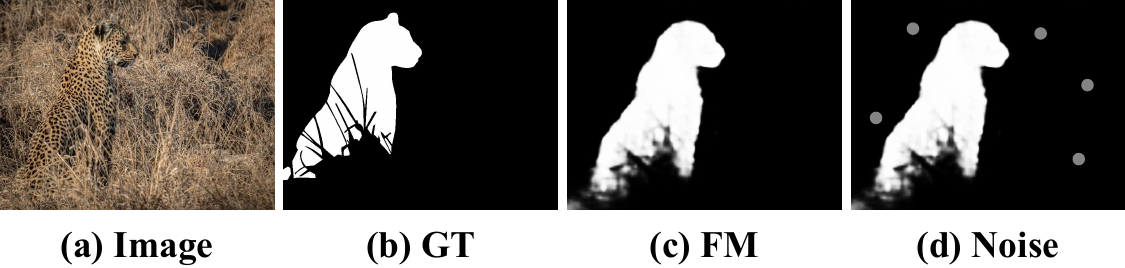}
    \caption{\textbf{Meta-measure 3: noise sensitivity.} 
    Adding slight random noise in FM will degrade its predicted quality. For ease of visualization, the noise in (d) is presented in an exaggerated form.
    }
    \label{fig:mm3}
\end{figure}

\subsection{Meta-Measure 3: Noise Sensitivity}\label{sec:mm3}

When slight random noise is introduced into a predicted FM, it may not significantly alter the structural content but can still degrade its visual quality, as shown in \figref{fig:mm3}. An ideal evaluation metric should be sensitive to such subtle changes and accurately reflect their impact on perceived quality. To assess this property, we design MM\#3 to evaluate the perceptual sensitivity of different metrics to fine-grained variations in the FMs.

\noindent\textbf{Evaluation Protocol.} \textit{(a) Image Selection.} To guarantee the quality of the predicted FMs under evaluation, we select high-quality predicted FMs (F1 score $\ge$ 0.6) from all experimental data. \textit{(b) Noise Addition.} We randomly sample pixels accounting for 1\% of the total image pixels within the intersection of the predicted FM and the GT background regions, where Gaussian noise $\epsilon \sim \CMcal{N}(0,0.2^2)$ is added (negative noise values will be truncated). \textit{(c) Error Recording.} We evaluate both the noisy and original predicted FMs against the GT, defining the error rate as the proportion of cases in which the noisy version outperforms the original. A lower error rate indicates that the metric is more sensitive to such fine-grained disturbances and thus more reliable in detecting subtle prediction degradation.

\noindent\textbf{Result Analysis.} As shown in \tabref{tab:result-mm}, columns 6,7,8, IoU, $F_{\beta}$ and $F_{\beta}^{\omega}$ all achieve excellent performance in this meta-measure, which is expected given their pixel-level sensitivity, enabling them to capture such subtle errors. 
Theoretically, IoU and $F_{\beta}$ should perform as well as $F_{\beta}^{\omega}$. However, this is not observed in practice. The key reason is that $F_{\beta}^{\omega}$ supports non-binary inputs, whereas $\text{IoU}, F_{\beta}$ rely on adaptive thresholding for binarization. When the noise is extremely slight, adaptive thresholding tends to classify such noise as background, yielding identical scores before and after perturbation and thus failing to detect the change.
Unlike these metrics, $S_{\alpha}$ exhibits significantly poorer performance on the experimental samples from COD10K and NC4K.
This is because $S_\alpha$ divides the image into regions and compares statistical information within each region. For predictions of moderate quality (\ie, from COD10K or NC4K), adding noise may shift regional statistics closer to the GT, occasionally resulting in higher scores. For high-quality predictions (\ie, from Trans10K), even minor noise can disrupt regional structures, allowing $S_\alpha$ to demonstrate a level of sensitivity to perturbations that may surpass that of IoU and $F_{\beta}$. 
Our $C_{\beta}^{\omega}$, based on the pixel correlation framework that captures relationships among arbitrary pixels, exhibits exceptional sensitivity to even the slightest noise perturbations, achieving superior performance across all datasets. 

\begin{figure}[t!]
    \centering
    \includegraphics[width=1\linewidth]{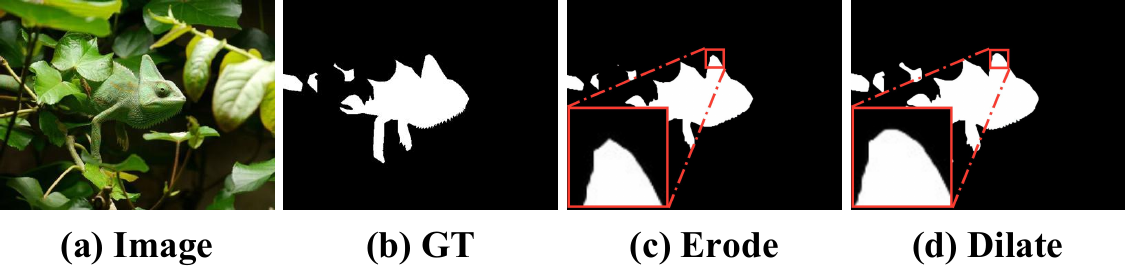}
    \caption{
    \textbf{Meta-measure 4: annotation boundary.}
    Mild morphological perturbations are applied to the GT to simulate the boundary uncertainty commonly introduced during manual annotation. 
    }
    \label{fig:mm4}
\end{figure}

\subsection{Meta-Measure 4: Annotation Boundary}\label{sec:mm4}

For highly camouflaged objects, accurately delineating their boundaries poses a significant challenge. Due to inherent uncertainty at the boundaries, manually annotated GT masks inevitably involve subjective judgments, which may result in misalignment with the actual object boundaries, especially in regions with weak local visual evidence. We argue that an ideal metric should be robust to such boundary deviations within a reasonable range, maintaining stability in its evaluation results. To this end, we propose the final meta-measure (MM\#4), which aims to assess the stability of each metric under conditions of boundary ambiguity. An illustrative example of this meta-measure is shown in \figref{fig:mm4}.

\noindent\textbf{Evaluation Protocol.} To simulate the variability between annotations, we applied mild morphological operations -- dilation and erosion -- to the GT mask, introducing subtle, non-destructive boundary perturbations. We proceed in two steps. \textit{(a) Boundary Perturbation.} A $3\times 3$ kernel is applied to perform dilation and erosion on the GT masks of all experimental samples. \textit{(b) Score Variation Recording.} The absolute score differences in each evaluation metric before and after the boundary perturbation are recorded and used as the results of this meta-measure study.

\noindent\textbf{Result Analysis.} As shown in \tabref{tab:result-mm}, columns 9--14, since only mild and subtle boundary perturbations are introduced, all metrics exhibit strong stability under this setting, with their average variations consistently maintained around the 0.01 level.
Additionally, the results show that the average variation after dilation tends to be smaller than that after erosion. Despite this, our $C_{\beta}^{\omega}$ consistently achieves the smallest variations, demonstrating superior robustness to boundary ambiguity -- a particularly desirable property in COS where annotation uncertainty is inherent.

\begin{table*}[t!]
    \centering
    \renewcommand{\arraystretch}{0.6}
    \setlength\tabcolsep{1.4pt}
    \caption{
    \textbf{Benchmark on camouflaged object segmentation task.}
    We comprehensively evaluate 40 representative COS methods spanning the development of the field from its earliest dedicated models to the latest available approaches, including both CNN- and Transformer-based architectures. All results are computed using the predicted maps released by the original authors to ensure a fair comparison across three widely adopted datasets.
    The best and second-best results are highlighted in \red{red} and \blue{blue}, respectively.
    }
    \label{tab:cod_benchmark}
    \begin{tabular}{lccccccc>{\columncolor{gray!20}}ccccc>{\columncolor{gray!20}}ccccc>{\columncolor{gray!20}}c}
\toprule
\multirow[c]{2}{*}[-1.0ex]{Methods} & \multirow[c]{2}{*}[-1.0ex]{Pub./Year} & \multirow[c]{2}{*}[-1.0ex]{Size} & \multirow[c]{2}{*}[-1.0ex]{Backbone} & \multicolumn{5}{c}{CAMO~\cite{le2019anabranch}} & \multicolumn{5}{c}{COD10K~\cite{fan2020camouflaged}} & \multicolumn{5}{c}{NC4K~\cite{lv2021simultaneously}}\\ \cmidrule(lr){5-9} \cmidrule(lr){10-14} \cmidrule(lr){15-19}
 &  &  &  & $\CMcal{M}$ & $F_{\beta}^{\omega}$ & $S_{\alpha}$ & $E_{\phi}$ & $C_{\beta}^{\omega}$ & $\CMcal{M}$ & $F_{\beta}^{\omega}$ & $S_{\alpha}$ & $E_{\phi}$ & $C_{\beta}^{\omega}$ & $\CMcal{M}$ & $F_{\beta}^{\omega}$ & $S_{\alpha}$ & $E_{\phi}$ & $C_{\beta}^{\omega}$ \\
\midrule
\multicolumn{19}{c}{CNN-Based Methods}\\
\midrule
\multirow{2}{*}{SINet~\cite{fan2020camouflaged}} & \multirow{2}{*}{CVPR$_{20}$} & \multirow{2}{*}{352 × 352} & \multirow{2}{*}{ResNet-50} & 0.100 & 0.606 & 0.751 & 0.834 & 0.650 & 0.051 & 0.551 & 0.771 & 0.797 & 0.618 & 0.058 & 0.723 & 0.808 & 0.883 & 0.740 \\
&&&&19 & 19 & 19 & 19 & 19 & 19 & 19 & 19 & 19 & 19 & 18 & 19 & 19 & 17 & 19 \\
\multirow{2}{*}{PraNet~\cite{fan2020pranet}} & \multirow{2}{*}{MICCAI$_{20}$} & \multirow{2}{*}{352 × 352} & \multirow{2}{*}{Res2Net-50} & 0.094 & 0.663 & 0.769 & 0.835 & 0.707 & 0.045 & 0.629 & 0.789 & 0.840 & 0.684 & 0.059 & 0.724 & 0.822 & 0.875 & 0.760 \\
&&&&18 & 18 & 18 & 18 & 17 & 18 & 18 & 18 & 18 & 17 & 19 & 18 & 18 & 19 & 17 \\
\multirow{2}{*}{TINet~\cite{zhu2021inferring}} & \multirow{2}{*}{AAAI$_{21}$} & \multirow{2}{*}{352 × 352} & \multirow{2}{*}{ResNet-50} & 0.087 & 0.678 & 0.781 & 0.847 & 0.723 & 0.042 & 0.635 & 0.793 & 0.848 & 0.684 & 0.055 & 0.734 & 0.829 & 0.882 & 0.761 \\
&&&&16 & 16 & 16 & 17 & 13 & 17 & 17 & 17 & 17 & 18 & 17 & 17 & 17 & 18 & 16 \\
\multirow{2}{*}{LSR~\cite{lv2021simultaneously}} & \multirow{2}{*}{CVPR$_{21}$} & \multirow{2}{*}{352 × 352} & \multirow{2}{*}{ResNet-50} & 0.080 & 0.696 & 0.787 & 0.859 & 0.722 & 0.037 & 0.673 & 0.804 & 0.883 & 0.703 & 0.048 & 0.766 & 0.840 & 0.904 & 0.776 \\
&&&&13 & 13 & 13 & 14 & 15 & 14 & 13 & 15 & 9 & 14 & 11 & 11 & 11 & 9 & 13 \\
\multirow{2}{*}{MGL-R~\cite{zhai2021mutual}} & \multirow{2}{*}{CVPR$_{21}$} & \multirow{2}{*}{473 × 473} & \multirow{2}{*}{ResNet-50} & 0.088 & 0.673 & 0.775 & 0.848 & 0.700 & 0.035 & 0.666 & 0.814 & 0.864 & 0.697 & 0.053 & 0.739 & 0.833 & 0.889 & 0.757 \\
&&&&17 & 17 & 17 & 16 & 18 & 9 & 15 & 12 & 14 & 16 & 15 & 16 & 15 & 15 & 18 \\
\multirow{2}{*}{PFNet~\cite{mei2021camouflaged}} & \multirow{2}{*}{CVPR$_{21}$} & \multirow{2}{*}{416 × 416} & \multirow{2}{*}{ResNet-50} & 0.085 & 0.695 & 0.782 & 0.855 & 0.723 & 0.040 & 0.660 & 0.800 & 0.868 & 0.700 & 0.053 & 0.745 & 0.829 & 0.894 & 0.766 \\
&&&&14 & 14 & 15 & 15 & 14 & 16 & 16 & 16 & 13 & 15 & 16 & 15 & 16 & 14 & 15 \\
\multirow{2}{*}{UJSCOD~\cite{li2021uncertainty}} & \multirow{2}{*}{CVPR$_{21}$} & \multirow{2}{*}{352 × 352} & \multirow{2}{*}{Res2Net-50} & 0.073 & 0.728 & 0.800 & 0.872 & 0.748 & 0.035 & 0.684 & 0.809 & 0.882 & 0.713 & 0.047 & 0.771 & 0.842 & 0.906 & 0.782 \\
&&&&9 & 10 & 11 & 11 & 11 & 10 & 10 & 14 & 10 & 12 & 8 & 8 & 9 & 7 & 10 \\
\multirow{2}{*}{UGTR~\cite{yang2021uncertainty}} & \multirow{2}{*}{ICCV$_{21}$} & \multirow{2}{*}{473 × 473} & \multirow{2}{*}{ResNet-50} & 0.086 & 0.686 & 0.785 & 0.861 & 0.719 & 0.035 & 0.667 & 0.818 & 0.850 & 0.705 & 0.052 & 0.747 & 0.839 & 0.889 & 0.770 \\
&&&&15 & 15 & 14 & 13 & 16 & 11 & 14 & 9 & 16 & 13 & 14 & 14 & 12 & 16 & 14 \\
\multirow{2}{*}{C2FNet~\cite{sun2021context}} & \multirow{2}{*}{IJCAI$_{21}$} & \multirow{2}{*}{352 × 352} & \multirow{2}{*}{Res2Net-50} & 0.080 & 0.719 & 0.796 & 0.865 & 0.744 & 0.036 & 0.686 & 0.813 & 0.886 & 0.722 & 0.049 & 0.762 & 0.838 & 0.901 & 0.784 \\
&&&&12 & 12 & 12 & 12 & 12 & 13 & 9 & 13 & 8 & 10 & 12 & 12 & 13 & 12 & 9 \\
\multirow{2}{*}{FDNet~\cite{zhong2022detecting}} & \multirow{2}{*}{CVPR$_{22}$} & \multirow{2}{*}{416 × 416} & \multirow{2}{*}{Res2Net-50} & \blue{0.063} & \red{0.775} & \red{0.842} & \blue{0.901} & \red{0.808} & 0.030 & \blue{0.729} & \red{0.840} & \blue{0.906} & \red{0.768} & 0.052 & 0.750 & 0.834 & 0.895 & 0.779 \\
&&&&2 & 1 & 1 & 2 & 1 & 3 & 2 & 1 & 2 & 1 & 13 & 13 & 14 & 13 & 12 \\
\multirow{2}{*}{SegMaR~\cite{jia2022segment}} & \multirow{2}{*}{CVPR$_{22}$} & \multirow{2}{*}{352 × 352} & \multirow{2}{*}{ResNet-50} & 0.071 & 0.753 & 0.815 & 0.881 & 0.771 & 0.034 & 0.724 & 0.833 & 0.893 & 0.745 & 0.046 & 0.781 & 0.841 & 0.905 & 0.779 \\
&&&&7 & 3 & 6 & 7 & 7 & 8 & 4 & 4 & 6 & 4 & 7 & 7 & 10 & 8 & 11 \\
\multirow{2}{*}{ZoomNet~\cite{pang2022zoom}} & \multirow{2}{*}{CVPR$_{22}$} & \multirow{2}{*}{384 × 384} & \multirow{2}{*}{ResNet-50} & 0.066 & 0.752 & 0.820 & 0.882 & 0.773 & \red{0.029} & 0.729 & 0.838 & 0.893 & 0.742 & 0.043 & 0.784 & 0.853 & 0.907 & 0.792 \\
&&&&3 & 4 & 5 & 6 & 5 & 1 & 3 & 3 & 7 & 6 & 3 & 5 & 4 & 6 & 7 \\
\multirow{2}{*}{BGNet~\cite{sun2022boundary}} & \multirow{2}{*}{IJCAI$_{22}$} & \multirow{2}{*}{352 × 352} & \multirow{2}{*}{Res2Net-50} & 0.073 & 0.749 & 0.812 & 0.876 & 0.770 & 0.033 & 0.722 & 0.831 & 0.902 & 0.745 & 0.044 & 0.788 & 0.851 & 0.911 & 0.799 \\
&&&&10 & 5 & 8 & 10 & 8 & 6 & 6 & 6 & 3 & 5 & 6 & 4 & 5 & 4 & 4 \\
\multirow{2}{*}{SINet-V2~\cite{fan2021concealed}} & \multirow{2}{*}{TPAMI$_{22}$} & \multirow{2}{*}{352 × 352} & \multirow{2}{*}{Res2Net-50} & 0.070 & 0.743 & 0.820 & 0.884 & 0.778 & 0.037 & 0.680 & 0.815 & 0.864 & 0.726 & 0.048 & 0.770 & 0.847 & 0.901 & 0.794 \\
&&&&6 & 8 & 4 & 3 & 3 & 15 & 12 & 11 & 15 & 9 & 10 & 9 & 6 & 11 & 6 \\
\multirow{2}{*}{TPRNet~\cite{zhang2023tprnet}} & \multirow{2}{*}{TVCJ$_{22}$} & \multirow{2}{*}{352 × 352} & \multirow{2}{*}{Res2Net-50} & 0.074 & 0.725 & 0.807 & 0.880 & 0.756 & 0.036 & 0.683 & 0.817 & 0.869 & 0.721 & 0.048 & 0.768 & 0.846 & 0.901 & 0.786 \\
&&&&11 & 11 & 9 & 8 & 10 & 12 & 11 & 10 & 12 & 11 & 9 & 10 & 8 & 10 & 8 \\
\multirow{2}{*}{FEDER~\cite{he2023camouflaged}} & \multirow{2}{*}{CVPR$_{23}$} & \multirow{2}{*}{384 × 384} & \multirow{2}{*}{ResNet-50} & 0.071 & 0.738 & 0.802 & 0.877 & 0.760 & 0.032 & 0.716 & 0.822 & 0.901 & 0.736 & 0.044 & 0.789 & 0.847 & \blue{0.913} & 0.794 \\
&&&&8 & 9 & 10 & 9 & 9 & 5 & 7 & 7 & 5 & 8 & 5 & 3 & 7 & 2 & 5 \\
\multirow{2}{*}{DGNet~\cite{ji2023deep}} & \multirow{2}{*}{MIR$_{23}$} & \multirow{2}{*}{352 × 352} & \multirow{2}{*}{EfficientNet-B4} & \red{0.057} & \blue{0.769} & \blue{0.839} & \red{0.906} & \blue{0.800} & 0.033 & 0.693 & 0.822 & 0.879 & 0.736 & \red{0.042} & 0.784 & \red{0.857} & 0.910 & \blue{0.805} \\
&&&&1 & 2 & 2 & 1 & 2 & 7 & 8 & 8 & 11 & 7 & 1 & 6 & 1 & 5 & 2 \\
\multirow{2}{*}{DINet~\cite{zhou2024decoupling}} & \multirow{2}{*}{TMM$_{24}$} & \multirow{2}{*}{400 × 400} & \multirow{2}{*}{Res2Net-50} & 0.068 & 0.748 & 0.821 & 0.884 & 0.778 & 0.031 & 0.724 & 0.832 & 0.902 & 0.750 & 0.043 & \red{0.790} & \blue{0.856} & 0.912 & 0.805 \\
&&&&4 & 6 & 3 & 4 & 4 & 4 & 5 & 5 & 4 & 3 & 4 & 1 & 2 & 3 & 3 \\
\multirow{2}{*}{RUN~\cite{he2025run}} & \multirow{2}{*}{ICML$_{25}$} & \multirow{2}{*}{352 × 352} & \multirow{2}{*}{ResNet-50} & 0.069 & 0.744 & 0.813 & 0.884 & 0.772 & \blue{0.029} & \red{0.735} & \blue{0.839} & \red{0.910} & \blue{0.767} & \blue{0.042} & \blue{0.790} & 0.854 & \red{0.916} & \red{0.808} \\
&&&&5 & 7 & 7 & 5 & 6 & 2 & 1 & 2 & 1 & 2 & 2 & 2 & 3 & 1 & 1 \\
\midrule
\multicolumn{19}{c}{Transformer-Based Methods}\\
\midrule
\multirow{2}{*}{DTINet~\cite{liu2022boosting}} & \multirow{2}{*}{ICPR$_{22}$} & \multirow{2}{*}{256 × 256} & \multirow{2}{*}{ViT-B} & 0.050 & 0.796 & 0.856 & 0.918 & 0.829 & 0.034 & 0.695 & 0.824 & 0.881 & 0.737 & 0.041 & 0.792 & 0.863 & 0.914 & 0.816 \\
&&&&20 & 21 & 20 & 21 & 20 & 21 & 21 & 21 & 21 & 21 & 21 & 21 & 21 & 21 & 21 \\
\multirow{2}{*}{FSPNet~\cite{huang2023feature}} & \multirow{2}{*}{CVPR$_{23}$} & \multirow{2}{*}{384 × 384} & \multirow{2}{*}{ViT-B} & 0.050 & 0.799 & 0.856 & 0.919 & 0.820 & 0.026 & 0.735 & 0.851 & 0.900 & 0.762 & 0.035 & 0.816 & 0.879 & 0.923 & 0.827 \\
&&&&19 & 20 & 21 & 20 & 21 & 20 & 20 & 20 & 20 & 20 & 19 & 20 & 20 & 20 & 20 \\
\multirow{2}{*}{RISNet~\cite{wang2024depth}} & \multirow{2}{*}{CVPR$_{24}$} & \multirow{2}{*}{704 × 704} & \multirow{2}{*}{PVTv2-B2} & 0.050 & 0.827 & 0.870 & 0.922 & 0.847 & 0.025 & 0.799 & 0.873 & 0.927 & 0.806 & 0.037 & 0.834 & 0.882 & 0.926 & 0.837 \\
&&&&21 & 17 & 16 & 18 & 15 & 18 & 10 & 11 & 14 & 11 & 20 & 19 & 18 & 19 & 19 \\
\multirow{2}{*}{VSCode~\cite{luo2024vscode}} & \multirow{2}{*}{CVPR$_{24}$} & \multirow{2}{*}{352 × 352} & \multirow{2}{*}{Swin-S} & 0.046 & 0.820 & 0.873 & 0.928 & 0.843 & 0.023 & 0.780 & 0.869 & 0.929 & 0.796 & 0.032 & 0.841 & 0.891 & 0.939 & 0.845 \\
&&&&15 & 18 & 14 & 13 & 18 & 14 & 17 & 16 & 12 & 17 & 13 & 17 & 12 & 12 & 15 \\
\multirow{2}{*}{FSEL~\cite{sun2024frequency}} & \multirow{2}{*}{ECCV$_{24}$} & \multirow{2}{*}{416 × 416} & \multirow{2}{*}{PVTv2-B4} & 0.040 & 0.851 & 0.885 & 0.942 & 0.867 & 0.021 & 0.800 & 0.877 & 0.928 & 0.815 & 0.030 & 0.853 & 0.893 & 0.941 & 0.855 \\
&&&&5 & 5 & 5 & 4 & 4 & 9 & 8 & 7 & 13 & 7 & 8 & 10 & 8 & 9 & 9 \\
\multirow{2}{*}{DPRNet~\cite{zha2024dual}} & \multirow{2}{*}{TCSVT$_{24}$} & \multirow{2}{*}{384 × 384} & \multirow{2}{*}{PVTv2-B2} & 0.046 & 0.830 & 0.865 & 0.931 & 0.845 & 0.025 & 0.771 & 0.854 & 0.926 & 0.788 & 0.033 & 0.838 & 0.880 & 0.937 & 0.840 \\
&&&&14 & 15 & 19 & 12 & 17 & 19 & 19 & 19 & 17 & 19 & 17 & 18 & 19 & 15 & 18 \\
\multirow{2}{*}{CamoFormer~\cite{yin2024camoformer}} & \multirow{2}{*}{TPAMI$_{24}$} & \multirow{2}{*}{384 × 384} & \multirow{2}{*}{PVTv2-B4} & 0.046 & 0.831 & 0.872 & 0.931 & 0.848 & 0.023 & 0.786 & 0.869 & 0.931 & 0.801 & 0.030 & 0.847 & 0.892 & 0.941 & 0.850 \\
&&&&16 & 14 & 15 & 11 & 14 & 12 & 16 & 15 & 11 & 15 & 9 & 12 & 10 & 8 & 11 \\
\multirow{2}{*}{ZoomNeXt~\cite{pang2024zoomnext}} & \multirow{2}{*}{TPAMI$_{24}$} & \multirow{2}{*}{384 × 384} & \multirow{2}{*}{PVTv2-B4} & 0.040 & \blue{0.859} & 0.888 & 0.935 & 0.868 & 0.017 & \blue{0.838} & \blue{0.898} & 0.945 & 0.836 & 0.028 & 0.865 & 0.900 & 0.943 & 0.860 \\
&&&&6 & 2 & 4 & 8 & 3 & 3 & 2 & 2 & 4 & 4 & 3 & 3 & 4 & 7 & 3 \\
\multirow{2}{*}{ESCNet~\cite{ye2025escnet}} & \multirow{2}{*}{ICCV$_{25}$} & \multirow{2}{*}{416 × 416} & \multirow{2}{*}{PVTv2-B5} & 0.041 & 0.849 & 0.875 & 0.940 & 0.858 & 0.020 & 0.808 & 0.873 & 0.944 & 0.813 & 0.028 & 0.864 & 0.893 & 0.949 & 0.858 \\
&&&&7 & 6 & 11 & 6 & 11 & 6 & 6 & 10 & 6 & 9 & 4 & 4 & 9 & 3 & 5 \\
\multirow{2}{*}{SAM-TTT~\cite{yu2025sam}} & \multirow{2}{*}{MM$_{25}$} & \multirow{2}{*}{1024 × 1024} & \multirow{2}{*}{SAM ViT-B} & 0.046 & 0.837 & 0.869 & 0.942 & 0.855 & 0.024 & 0.801 & 0.874 & 0.932 & 0.819 & 0.033 & 0.854 & 0.890 & 0.946 & 0.857 \\
&&&&13 & 12 & 17 & 5 & 12 & 17 & 7 & 9 & 10 & 5 & 18 & 9 & 14 & 5 & 7 \\
\multirow{2}{*}{CFRN~\cite{song2025continuous}} & \multirow{2}{*}{TIP$_{25}$} & \multirow{2}{*}{384 × 384} & \multirow{2}{*}{Swin-B} & 0.039 & 0.849 & 0.881 & 0.943 & 0.864 & 0.021 & 0.800 & 0.872 & 0.939 & 0.812 & 0.029 & 0.855 & 0.891 & 0.947 & 0.856 \\
&&&&4 & 7 & 9 & 3 & 6 & 7 & 9 & 12 & 7 & 10 & 6 & 8 & 11 & 4 & 8 \\
\multirow{2}{*}{SENet~\cite{hao2025simple}} & \multirow{2}{*}{TIP$_{25}$} & \multirow{2}{*}{384 × 384} & \multirow{2}{*}{MAE} & 0.039 & 0.847 & 0.888 & 0.927 & 0.864 & 0.024 & 0.779 & 0.865 & 0.919 & 0.795 & 0.032 & 0.843 & 0.889 & 0.930 & 0.845 \\
&&&&3 & 9 & 3 & 14 & 5 & 16 & 18 & 18 & 19 & 18 & 15 & 16 & 16 & 18 & 17 \\
\multirow{2}{*}{CODIB~\cite{li2026learning}} & \multirow{2}{*}{TMM$_{25}$} & \multirow{2}{*}{384 × 384} & \multirow{2}{*}{PVTv2-B2} & 0.045 & 0.831 & 0.875 & 0.924 & 0.850 & 0.023 & 0.787 & 0.870 & 0.926 & 0.802 & 0.032 & 0.845 & 0.890 & 0.932 & 0.847 \\
&&&&12 & 13 & 12 & 17 & 13 & 13 & 14 & 14 & 16 & 13 & 14 & 15 & 15 & 17 & 13 \\
\multirow{2}{*}{SFCNet~\cite{zhao2025spatial}} & \multirow{2}{*}{TMM$_{25}$} & \multirow{2}{*}{384 × 384} & \multirow{2}{*}{SMT-T} & 0.042 & 0.846 & 0.882 & 0.934 & 0.860 & 0.022 & 0.797 & 0.872 & 0.938 & 0.805 & 0.031 & 0.850 & 0.891 & 0.939 & 0.847 \\
&&&&9 & 10 & 7 & 9 & 8 & 10 & 12 & 13 & 8 & 12 & 11 & 11 & 13 & 11 & 14 \\
\multirow{2}{*}{UTNet~\cite{sun2025unet}} & \multirow{2}{*}{TMM$_{25}$} & \multirow{2}{*}{384 × 384} & \multirow{2}{*}{SMT-T} & 0.049 & 0.829 & 0.868 & 0.926 & 0.845 & 0.022 & 0.791 & 0.868 & 0.933 & 0.799 & 0.032 & 0.846 & 0.887 & 0.938 & 0.845 \\
&&&&18 & 16 & 18 & 16 & 16 & 11 & 13 & 17 & 9 & 16 & 16 & 13 & 17 & 13 & 16 \\
\multirow{2}{*}{Camodiffusion~\cite{sun2025conditional}} & \multirow{2}{*}{TPAMI$_{25}$} & \multirow{2}{*}{384 × 384} & \multirow{2}{*}{PVTv2-B4} & 0.042 & 0.851 & 0.878 & 0.940 & 0.860 & 0.019 & 0.817 & 0.883 & 0.946 & 0.818 & 0.028 & 0.861 & 0.895 & 0.946 & 0.854 \\
&&&&10 & 4 & 10 & 7 & 9 & 4 & 5 & 5 & 3 & 6 & 5 & 5 & 7 & 6 & 10 \\
\multirow{2}{*}{VSCode-V2~\cite{luo2025vscode}} & \multirow{2}{*}{TPAMI$_{25}$} & \multirow{2}{*}{352 × 352} & \multirow{2}{*}{Swin-S} & 0.049 & 0.817 & 0.875 & 0.921 & 0.840 & 0.024 & 0.787 & 0.874 & 0.925 & 0.801 & 0.031 & 0.845 & 0.897 & 0.938 & 0.849 \\
&&&&17 & 19 & 13 & 19 & 19 & 15 & 15 & 8 & 18 & 14 & 12 & 14 & 6 & 14 & 12 \\
\multirow{2}{*}{GBNet~\cite{wang2026gbnet}} & \multirow{2}{*}{TIP$_{26}$} & \multirow{2}{*}{704 × 704} & \multirow{2}{*}{PVTv2-B4} & 0.044 & 0.847 & 0.881 & 0.926 & 0.862 & 0.019 & 0.837 & 0.898 & 0.945 & \blue{0.838} & 0.030 & 0.860 & 0.897 & 0.935 & 0.858 \\
&&&&11 & 8 & 8 & 15 & 7 & 5 & 3 & 3 & 5 & 2 & 10 & 6 & 5 & 16 & 4 \\
\multirow{2}{*}{ICL-Camo~\cite{chen2026learn}} & \multirow{2}{*}{TIP$_{26}$} & \multirow{2}{*}{392 × 392} & \multirow{2}{*}{ViT-B} & \blue{0.037} & 0.859 & \blue{0.892} & \blue{0.944} & \blue{0.870} & \blue{0.017} & 0.834 & 0.896 & \blue{0.950} & 0.837 & \blue{0.024} & \blue{0.879} & \blue{0.912} & \blue{0.954} & \blue{0.873} \\
&&&&2 & 3 & 2 & 2 & 2 & 2 & 4 & 4 & 2 & 3 & 2 & 2 & 2 & 2 & 2 \\
\multirow{2}{*}{SAM2-UNet~\cite{xiong2026sam2}} & \multirow{2}{*}{VINT$_{26}$} & \multirow{2}{*}{352 × 352} & \multirow{2}{*}{SAM2-Hiera-L} & 0.042 & 0.845 & 0.884 & 0.933 & 0.859 & 0.021 & 0.798 & 0.880 & 0.926 & 0.813 & 0.029 & 0.856 & 0.901 & 0.940 & 0.857 \\
&&&&8 & 11 & 6 & 10 & 10 & 8 & 11 & 6 & 15 & 8 & 7 & 7 & 3 & 10 & 6 \\
\multirow{2}{*}{DepthSAM~\cite{han2026beyond}} & \multirow{2}{*}{CVPR$_{26}$} & \multirow{2}{*}{512 × 512} & \multirow{2}{*}{DAv2 ViT-L} & \red{0.028} & \red{0.895} & \red{0.919} & \red{0.960} & \red{0.898} & \red{0.014} & \red{0.872} & \red{0.920} & \red{0.960} & \red{0.864} & \red{0.021} & \red{0.902} & \red{0.929} & \red{0.962} & \red{0.890} \\
&&&&1 & 1 & 1 & 1 & 1 & 1 & 1 & 1 & 1 & 1 & 1 & 1 & 1 & 1 & 1 \\
\bottomrule
\end{tabular}
\end{table*}

\section{Generalization Analysis}\label{sec:generalization}

\subsection{Benchmark}

To further examine the practical applicability of the proposed Context-measure, we conduct benchmarks on COS and two application-oriented extensions of the COS task. COS serves as the primary benchmark, as it directly corresponds to the core motivation of this work. Polyp segmentation and mirror segmentation are further included as complementary application benchmarks to examine the applicability of Context-measure to specialized segmentation scenarios closely related to COS.

Across the three benchmarks, we compare the proposed $C_{\beta}^{\omega}$ with widely used segmentation metrics, including $\CMcal{M}$, $F_{\beta}^{\omega}$, $S_{\alpha}$, and $E_{\phi}$. For the COS benchmark, we additionally report the ranking of each model under every dataset-metric pair. Specifically, given a dataset and a metric, all models are sorted according to their corresponding metric values, and the rank is reported below the score, where rank $1$ denotes the best-performing model. 

\subsubsection{Camouflaged Object Segmentation}

As reported in \tabref{tab:cod_benchmark}, we evaluate 40 representative COS methods, covering both CNN- and Transformer-based architectures, on CAMO~\cite{le2019anabranch}, COD10K~\cite{fan2020camouflaged}, and NC4K~\cite{lv2021simultaneously}, which constitute the three standard benchmarks widely adopted in the COS field.
To ensure a fair comparison, all reported results are computed directly from the predicted maps released by the original authors.

\begin{figure}[t!]
    \centering
    \includegraphics[width=1\linewidth]{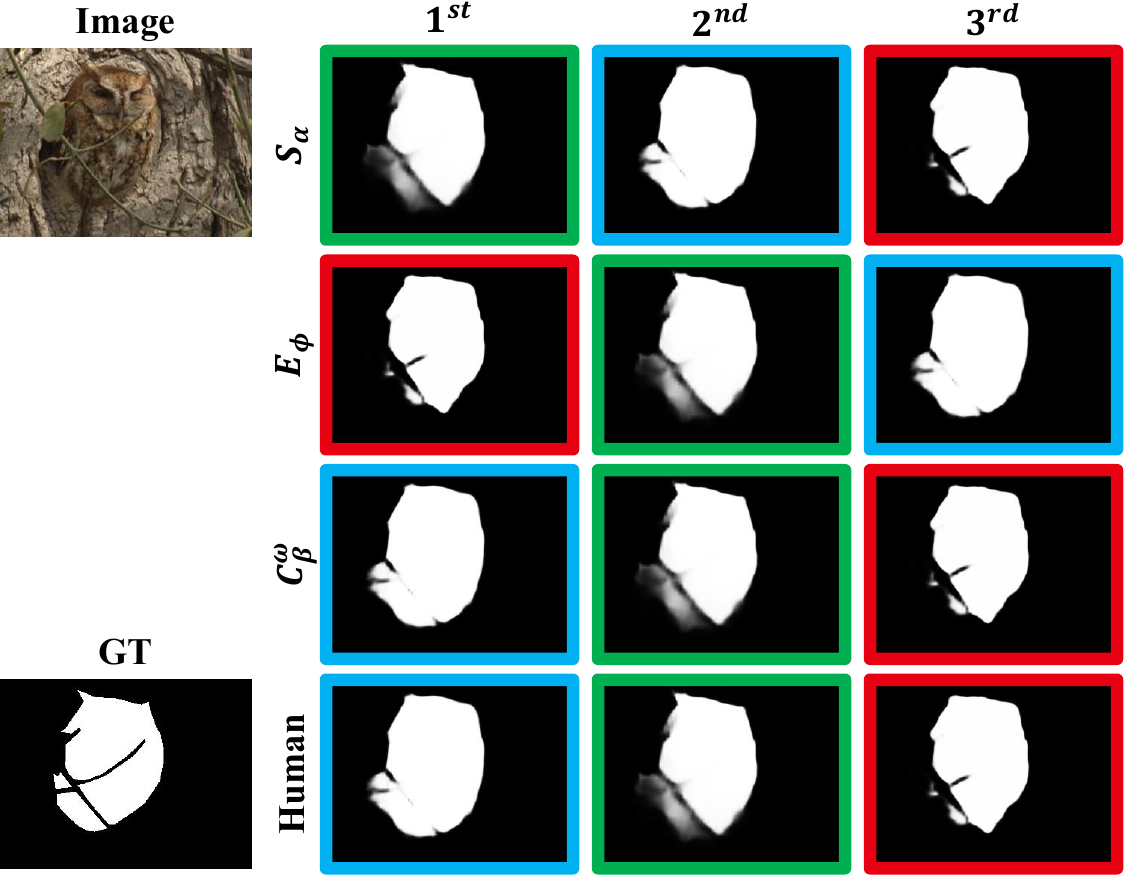}
    \caption{
    \textbf{Qualitative ranking results using different measures.}
    The blue-, green-, and red-bordered maps correspond to the FMs produced by FSEL~\cite{sun2024frequency}, TINet~\cite{zhu2021inferring}, and BGNet~\cite{sun2022boundary}, respectively. Compared with $S_{\alpha}$ and $E_{\phi}$, our $C_{\beta}^{\omega}$ produces a ranking that is more consistent with the human ranking shown at the bottom.
    }
    \label{fig:benchmark_demo}
\end{figure}

Across CAMO, COD10K, and NC4K, the rankings produced by $C_{\beta}^{\omega}$ remain highly correlated with those induced by existing metrics. Specifically, the average Spearman correlation between $C_{\beta}^{\omega}$ and $\{\CMcal{M},F_{\beta}^{\omega},S_{\alpha},E_{\phi}\}$ reaches $0.976$, and
the average Kendall's $\tau$ reaches $0.888$. The correlations are
consistent across datasets, with mean Spearman values of $0.980$,
$0.972$, and $0.975$ on CAMO, COD10K, and NC4K, respectively. This indicates that Context-measure preserves the broad performance consensus captured by existing metrics, especially for models with clear quality differences.
Meanwhile, high overall correlation does not mean that
$C_{\beta}^{\omega}$ is redundant. Although the dominant ranking trend is preserved, Context-measure still introduces meaningful ranking changes among closely performing models. The average pairwise inversion rate between $C_{\beta}^{\omega}$ and the widely used metrics is $5.6\%$, suggesting that a non-negligible subset of model pairs receive different relative orders under context-aware evaluation. An image-level example is shown in
\figref{fig:benchmark_demo}. In addition, the mean absolute shift between the conventional-metric consensus rank and the $C_{\beta}^{\omega}$ rank is $1.34$ positions. Some models show more evident changes. For example, SAM-TTT~\cite{yu2025sam} improves by $5.75$ positions on COD10K under $C_{\beta}^{\omega}$, while GBNet~\cite{wang2026gbnet} improves by $5.25$ positions on NC4K. Conversely, CamoDiffusion~\cite{sun2025conditional} drops by $4.25$ positions on NC4K. These changes suggest that Context-measure provides complementary discrimination rather than merely reproducing existing metric rankings.

We further inspect close competitors, where the benefit of
context-aware evaluation is most evident. We identify model pairs that are close under the conventional metrics but are more clearly separated by $C_{\beta}^{\omega}$. For instance, on CAMO, MGL-R~\cite{zhai2021mutual} and TINet~\cite{zhu2021inferring} have nearly identical conventional consensus ranks, differing by only $0.50$ positions, yet their $C_{\beta}^{\omega}$ ranks differ by $5$ positions. On COD10K, MGL-R and SINet-V2~\cite{fan2021concealed} differ by only $0.75$ positions under the conventional consensus rank, but differ by $7$ positions under $C_{\beta}^{\omega}$. These cases indicate that Context-measure is particularly useful for refining comparisons among models that appear nearly tied when evaluated only by context-blind metrics.

The above observations are consistent across model families. Within
CNN-based methods, $C_{\beta}^{\omega}$ maintains an average Spearman correlation of $0.932$ with conventional metrics, while still producing an average inversion rate of $10.1\%$. Within Transformer-based methods, the corresponding values are $0.887$ and $11.9\%$. Therefore, the additional discrimination introduced by $C_{\beta}^{\omega}$ is not confined to a specific architecture category. Instead, it reflects a general property of context-aware evaluation in COS.

Overall, the COS benchmark shows the practical value of
Context-measure for real model comparison. 
The proposed metric preserves the established performance consensus while providing additional discrimination among close competitors by accounting for contextual ambiguity, a defining characteristic of camouflaged scenarios.
Together with the meta-measure, these benchmark results show that Context-measure offers a more informative COS-oriented evaluation perspective for future model development.

\begin{table*}[t!]
    \centering
    \caption{
    \textbf{Benchmark on polyp segmentation task.}
    The best and second-best results are highlighted in \red{red} and \blue{blue}, respectively.
    }
    \label{tab:polyp_benchmark}
    \renewcommand{\arraystretch}{0.8}
    \setlength\tabcolsep{1.35pt}
    \begin{tabular}{lrcccc>{\columncolor{gray!20}}ccccc>{\columncolor{gray!20}}ccccc>{\columncolor{gray!20}}ccccc>{\columncolor{gray!20}}ccccc>{\columncolor{gray!20}}c}
    \toprule
    \multirow[c]{2}{*}[-0.9ex]{Methods} & \multicolumn{1}{c}{\multirow[c]{2}{*}[-0.9ex]{Pub./Year}} & \multicolumn{5}{c}{CVC-300~\cite{vazquez2017benchmark}} & \multicolumn{5}{c}{CVC-ClinicDB~\cite{bernal2015wm}} & \multicolumn{5}{c}{CVC-ColonDB~\cite{tajbakhsh2015automated}} & \multicolumn{5}{c}{ETIS~\cite{silva2014toward}} & \multicolumn{5}{c}{Kvasir~\cite{jha2019kvasir}} \\
    \cmidrule(lr){3-7} \cmidrule(lr){8-12} \cmidrule(lr){13-17} \cmidrule(lr){18-22} \cmidrule(lr){23-27}
     &  & $\CMcal{M}$ & $F_{\beta}^{\omega}$ & $S_{\alpha}$ & $E_{\phi}$ & $C_{\beta}^{\omega}$ & $\CMcal{M}$ & $F_{\beta}^{\omega}$ & $S_{\alpha}$ & $E_{\phi}$ & $C_{\beta}^{\omega}$ & $\CMcal{M}$ & $F_{\beta}^{\omega}$ & $S_{\alpha}$ & $E_{\phi}$ & $C_{\beta}^{\omega}$ & $\CMcal{M}$ & $F_{\beta}^{\omega}$ & $S_{\alpha}$ & $E_{\phi}$ & $C_{\beta}^{\omega}$ & $\CMcal{M}$ & $F_{\beta}^{\omega}$ & $S_{\alpha}$ & $E_{\phi}$ & $C_{\beta}^{\omega}$ \\
    \midrule
    UNet~\cite{ronneberger2015u} & MICCAI$_{15}$ & .022 & .684 & .843 & .867 & .717 & .019 & .811 & .889 & .917 & .804 & .059 & .491 & .710 & .758 & .515 & .036 & .366 & .684 & .645 & .418 & .055 & .794 & .858 & .901 & .818 \\
    SFA~\cite{fang2019selective} & MICCAI$_{19}$ & .065 & .341 & .640 & .604 & .478 & .042 & .647 & .793 & .816 & .686 & .094 & .366 & .629 & .634 & .462 & .109 & .231 & .557 & .515 & .305 & .075 & .670 & .782 & .828 & .724 \\
    UNet++~\cite{zhou2019unet++} & TMI$_{19}$ & .018 & .687 & .839 & .884 & .717 & .022 & .785 & .873 & .898 & .777 & .061 & .467 & .693 & .759 & .495 & .035 & .390 & .683 & .704 & .427 & .048 & .808 & .862 & .907 & .818 \\
    ACSNet~\cite{zhang2020adaptive} & MICCAI$_{20}$ & .013 & .825 & .923 & .916 & .854 & .011 & .873 & .927 & .955 & .857 & .039 & .697 & .829 & .861 & .715 & .059 & .530 & .754 & .774 & .582 & .032 & .882 & .920 & .944 & .897 \\
    PraNet~\cite{fan2020pranet} & MICCAI$_{20}$ & .010 & .843 & .925 & .938 & .864 & .009 & .896 & .936 & .957 & .872 & .043 & .699 & .820 & .847 & .713 & .031 & .600 & .794 & .792 & .634 & .030 & .885 & .915 & .943 & .892 \\
    EU-Net~\cite{patel2021enhanced} & CRV$_{21}$ & .015 & .805 & .904 & .915 & .831 & .011 & .891 & .936 & .959 & .876 & .045 & .730 & .831 & .867 & .754 & .067 & .636 & .793 & .807 & .693 & .028 & .893 & .917 & .945 & .901 \\
    SANet~\cite{wei2021shallow} & MICCAI$_{21}$ & .008 & .859 & .928 & .948 & .881 & .012 & .909 & .939 & .963 & .887 & .043 & .726 & .837 & .855 & .748 & .015 & .685 & .849 & .835 & .757 & .028 & .892 & .915 & .950 & .898 \\
    UACANet~\cite{kim2021uacanet} & MM$_{21}$ & \red{.005} & \red{.901} & .938 & \red{.980} & \red{.900} & \red{.006} & .928 & .942 & .976 & .895 & .039 & .746 & .835 & .878 & .746 & \red{.012} & .740 & .859 & \red{.905} & .771 & .025 & .902 & .917 & \blue{.958} & .905 \\
    PraNet-V2~\cite{hu2026pranet} & CVMJ$_{22}$ & .007 & \blue{.885} & \blue{.939} & .971 & .891 & .008 & \blue{.928} & .943 & \blue{.984} & .899 & .036 & .752 & .845 & .893 & .763 & .015 & .730 & .865 & .889 & .768 & .023 & .904 & .925 & .956 & .908 \\
    PolypPVT~\cite{dong2023polyp} & AIR$_{23}$ & .007 & .884 & .935 & \blue{.973} & \blue{.891} & \blue{.006} & \red{.936} & \blue{.949} & \red{.986} & \red{.904} & .031 & \red{.795} & \blue{.865} & \red{.919} & \blue{.802} & \blue{.013} & \blue{.750} & \blue{.871} & \blue{.905} & \blue{.791} & \blue{.023} & \blue{.911} & \blue{.925} & \red{.960} & \blue{.910} \\
    DGNet~\cite{ji2023deep} & MIR$_{23}$ & \blue{.006} & .880 & .939 & .963 & .891 & .009 & .898 & .933 & .969 & .875 & \blue{.030} & .765 & .858 & .891 & .780 & .018 & .690 & .847 & .870 & .740 & .030 & .887 & .910 & .950 & .891 \\
    CFANet~\cite{zhou2023cross} & PR$_{23}$ & .008 & .875 & .938 & .956 & .884 & .007 & .924 & \red{.950} & .971 & \blue{.900} & .039 & .728 & .835 & .887 & .744 & .014 & .693 & .845 & .872 & .741 & .023 & .903 & .924 & .950 & .908 \\
    SAM2-UNet~\cite{xiong2026sam2} & VINT$_{26}$ & .007 & .873 & \red{.940} & .954 & .883 & .009 & .900 & .946 & .961 & .879 & \red{.028} & \blue{.790} & \red{.877} & \blue{.909} & \red{.803} & .018 & \red{.759} & \red{.881} & .881 & \red{.800} & \red{.019} & \red{.919} & \red{.939} & .957 & \red{.921} \\
    \bottomrule
    \end{tabular}
\end{table*}

\subsubsection{Polyp Segmentation}

\begin{figure}[t!]
    \centering
    \includegraphics[width=1\linewidth]{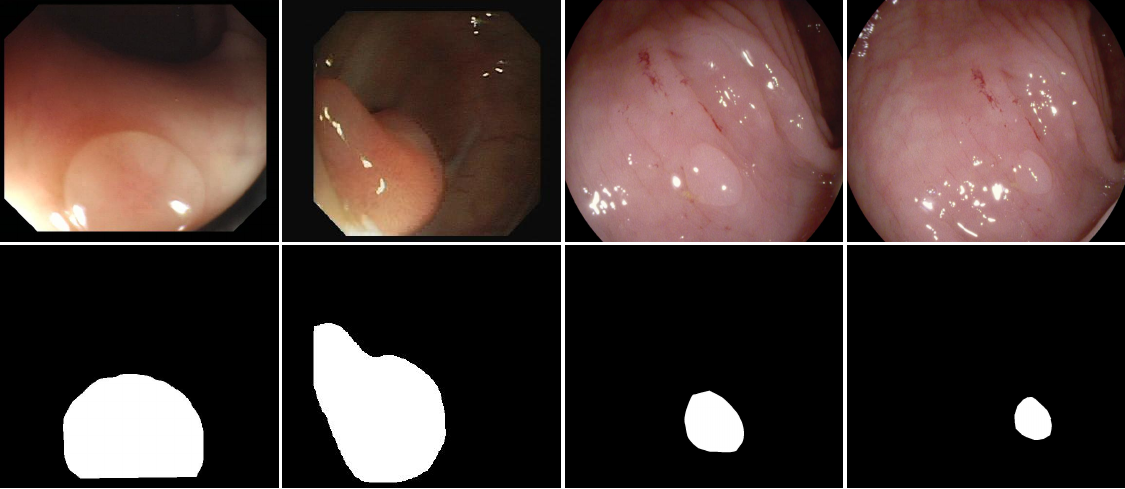}
    \caption{
    \textbf{Polyp segmentation.} The first row is input polyp images, while the second row shows their corresponding ground truths.
    }
    \label{fig:polyp}
\end{figure}

For the polyp segmentation task, we evaluate representative polyp segmentation models on five commonly used datasets, including CVC-300~\cite{vazquez2017benchmark}, CVC-ClinicDB~\cite{bernal2015wm}, CVC-ColonDB~\cite{tajbakhsh2015automated}, ETIS~\cite{silva2014toward}, and Kvasir~\cite{jha2019kvasir}. 
Unlike typical camouflaged objects, polyps do not necessarily blend into complex and varied backgrounds.
However, polyp regions often exhibit weak boundaries, low contrast, and high appearance similarity to surrounding mucosal tissues, making their segmentation quality closely related to local contextual cues, as shown in \figref{fig:polyp}.
These camouflage-like visual characteristics make polyp segmentation a meaningful medical application scenario for examining the broader applicability of the proposed COS-oriented evaluation metric.

As reported in \tabref{tab:polyp_benchmark}, the rankings produced by $C_{\beta}^{\omega}$ remain broadly consistent with those of conventional segmentation metrics. Across the five polyp datasets and four conventional metrics, the average Spearman correlation between $C_{\beta}^{\omega}$ and $\{\CMcal{M},F_{\beta}^{\omega},S_{\alpha},E_{\phi}\}$ is 0.934, and the average Kendall's $\tau$ is 0.840. The agreement is stable across datasets, with mean Spearman correlations of 0.940, 0.929, 0.920, 0.933, and 0.948 on CVC-300, CVC-ClinicDB, CVC-ColonDB, ETIS, and Kvasir, respectively, demonstrating that Context-measure preserves the overall performance trend of polyp segmentation models.
Meanwhile, $C_{\beta}^{\omega}$ is not a redundant transformation of existing metrics. The average pairwise inversion rate between Context-measure and the conventional metrics is 8.0\%, while the mean absolute shift from the conventional consensus rank is 0.64 positions. 
The shift remains relatively moderate overall, as clearly reflected by these aggregate ranking statistics, which is expected because polyp segmentation is less directly defined by camouflage than COS.
However, the effect becomes more visible on challenging datasets with weak boundaries and stronger appearance ambiguity. For example, on CVC-ColonDB, EU-Net~\cite{patel2021enhanced} improves by 2.75 positions under $C_{\beta}^{\omega}$ compared with the conventional consensus rank, whereas CFANet~\cite{zhou2023cross} drops by 2.25 positions; on ETIS, SAM2-UNet~\cite{xiong2026sam2} rises by 2.00 positions. We also observe this complementary behavior among close competitors. On CVC-ColonDB, ACSNet~\cite{zhang2020adaptive} and EU-Net differ by only 0.50 positions under the conventional consensus rank but by 4 positions under $C_{\beta}^{\omega}$, while PraNet~\cite{fan2020pranet} and SANet~\cite{wei2021shallow} show a corresponding gap increase from 1.50 to 4 positions. These results indicate that $C_{\beta}^{\omega}$ provides additional discrimination when conventional metrics assign similar scores to predicted FMs that differ in how well they segment visually ambiguous regions. Therefore, our Context-measure is also applicable to the evaluation of polyp segmentation.

\begin{figure}[t!]
    \centering
    \includegraphics[width=1\linewidth]{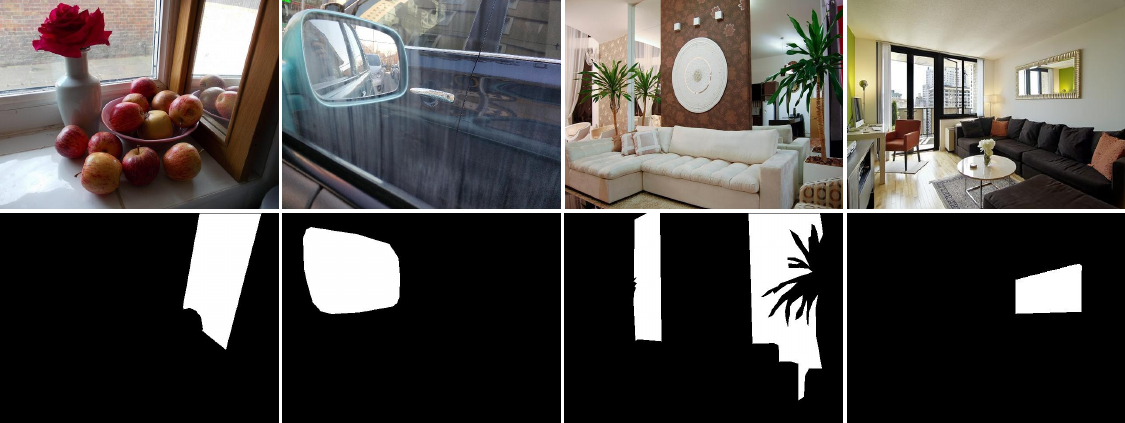}
    \caption{
    \textbf{Mirror segmentation.} The first row is mirror images, while the second row shows their corresponding ground truths.
    }
    \label{fig:mirror}
\end{figure}

\begin{table}[b!]
    \centering
    \renewcommand{\arraystretch}{0.8}
    \setlength\tabcolsep{1.05pt}
    \caption{
    \textbf{Benchmark on mirror segmentation task.}
    The best results are highlighted in \textbf{bold}.
    }
    \label{tab:mirror_benchmark}
    \begin{tabular}{lrcccc>{\columncolor{gray!20}}ccccc>{\columncolor{gray!20}}c}
    \toprule
\multirow[c]{2}{*}[-0.9ex]{Methods} & \multirow[c]{2}{*}[-0.9ex]{Pub./Year} & \multicolumn{5}{c}{MSD~\cite{yang2019my}} & \multicolumn{5}{c}{PMD~\cite{lin2020progressive}}\\
\cmidrule(lr){3-7} \cmidrule(lr){8-12}
    && $\CMcal{M}$ & $F_{\beta}^{\omega}$ & $S_{\alpha}$ & $E_{\phi}$ & $C_{\beta}^{\omega}$ & $\CMcal{M}$ & $F_{\beta}^{\omega}$ & $S_{\alpha}$ & $E_{\phi}$ & $C_{\beta}^{\omega}$\\
\midrule
MirrorNet~\cite{yang2019my} & ICCV$_{19}$ & .065 & .812 & .850 & .865 & .849 & .043 & .663 & .761 & .844 & .682 \\
PMD~\cite{lin2020progressive} & CVPR$_{20}$ & .047 & .845 & .875 & .913 & .857 & .032 & .716 & .810 & .866 & .732 \\
SANet~\cite{guan2022learning} & CVPR$_{22}$ & .054 & .829 & .862 & .903 & .843 & .071 & .721 & .808 & .839 & .732 \\
HetNet~\cite{he2023efficient} & AAAI$_{23}$ & .043 & .858 & .884 & .926 & .876 & .029 & .734 & .828 & .864 & .746 \\
DPRNet~\cite{zha2024dual} & TCSVT$_{24}$ & .033 & .888 & .904 & .938 & .893 & .027 & .766 & .844 & \textbf{.897} & .776 \\
MirrorSAM~\cite{zha2026seeing} & AAAI$_{26}$ & \textbf{.025} & \textbf{.924} & \textbf{.936} & \textbf{.953} & \textbf{.930} & \textbf{.024} & \textbf{.788} & \textbf{.868} & .894 & \textbf{.803} \\
\bottomrule
    \end{tabular}
\end{table}

\subsubsection{Mirror Segmentation}

For mirror segmentation, we conduct experiments on MSD~\cite{yang2019my} and PMD~\cite{lin2020progressive}. Mirror regions are often difficult to distinguish from their surroundings because their appearance is largely determined by reflected scenes rather than intrinsic object properties, as shown in \figref{fig:mirror}. Therefore, mirror segmentation provides another application-oriented setting for examining whether Context-measure can be applied beyond typical camouflage.

As reported in \tabref{tab:mirror_benchmark}, $C_{\beta}^{\omega}$ shows highly consistent rankings with conventional metrics on both mirror segmentation datasets. Across MSD and PMD, the average Spearman correlation between $C_{\beta}^{\omega}$ and $\{\CMcal{M},F_{\beta}^{\omega},S_{\alpha},E_{\phi}\}$ is $0.936$, and the average pairwise inversion rate is only $7.5\%$. The top-ranked models are also largely consistent across metrics, with average top-1 and top-3 overlaps of $0.875$ and $0.958$, respectively. In addition, the rank shifts induced by $C_{\beta}^{\omega}$ are small, with mean absolute shifts of $0.333$ on MSD and $0.250$ on PMD. These results indicate that Context-measure remains stable when applied to mirror segmentation, while preserving the overall model comparison trend of established metrics.

\begin{figure}[b!]
    \centering
    \includegraphics[width=1\linewidth]{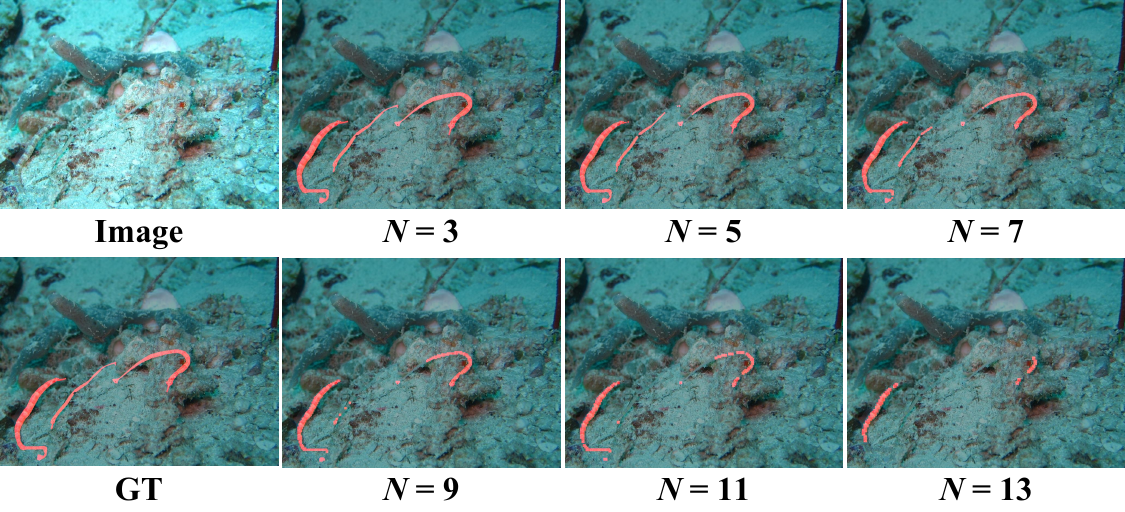}
    \caption{
    \textbf{Effect of patch size $N$ on target coverage.}
    As $N$ increases, the valid reconstruction region shrinks progressively, particularly around thin structures and object boundaries.
    }
    \label{fig:coverage}
\end{figure}

\section{Ablation Study}\label{sec:ablation}

We conduct an ablation study to examine both the hyperparameter configuration and the contribution of the key components in Context-measure. 
First, we analyze two groups of hyperparameters. The first group controls contextual affinity estimation, including the patch size $N$, band width $W$ (\secref{sec:n_w}), nonlinearity coefficient $\gamma$, and spatial weighting factor $\lambda$ (\secref{sec:gamma_lambda}). The second group governs the perception cycle and consists of the scale factor $\beta$ (\secref{sec:beta}). 
We then investigate two core design choices: the contribution of contextual affinity $\mD$ (\secref{sec:ablation_affinity}) and the effectiveness and generality of the pixel correlation framework (\secref{sec:ablation_framework}).
Finally, we examine the influence of color space on contextual affinity estimation (\secref{sec:ablation_colorspace}). Together, these experiments validate the default configuration and clarify the contribution of each major design.

\subsection{Hyperparameter Analysis}\label{sec:param}

\subsubsection{Patch Size \texorpdfstring{$N$}{N} and Band Width \texorpdfstring{$W$}{W}}
\label{sec:n_w}

The patch size $N$ and band width $W$ jointly determine the spatial configuration of contextual comparison. Specifically, $N$ controls the spatial support of each patch, whereas $W$ determines the range within which candidate patches are searched. Increasing $N$ allows each patch to capture richer structural context, but reduces target coverage because patches centered near object boundaries are more likely to extend beyond the target region, leaving some target pixels uncovered by any valid patch, as shown in \figref{fig:coverage}. To quantify this effect, we define the target coverage ratio $S_c$ as the proportion of target pixels covered by at least one valid patch and report its average over the COD10K test set under different values of $N$. As shown in \tabref{tab:s_ps}, $S_c$ consistently decreases as $N$ increases, dropping from 99.15\% at $N=3$ to 74.50\% at $N=13$. The largest reductions occur at smaller patch sizes, whereas the marginal decrease becomes progressively less pronounced as $N$ increases. Notably, $N=7$ still preserves an average target coverage of 87.53\%, providing a reasonable balance between spatial context and target coverage.
\begin{table}[t!]
    \centering
    \caption{\textbf{Target coverage.} Coverage $S_c$ decreases consistently as $N$ increases, averaged on COD10K test set.}
    \setlength\tabcolsep{6.65pt}
    \begin{tabular}{ccccccc}
        \toprule
        & $N=3$ & $N=5$ & $N=7$ & $N=9$ & $N=11$ & $N=13$ \\ \midrule
        $S_c$ & 99.15\% & 92.99\% & 87.53\% & 82.68\% & 78.44\% & 74.50\% \\ \bottomrule
    \end{tabular}
    \label{tab:s_ps}
\end{table}

\begin{table}[t!]
    \centering
    \caption{\textbf{Runtime under different combinations of patch size $N$ and band width $W$.} Results are averaged over 50 runs at two typical image resolutions. Runtime is primarily dominated by $N$: smaller values lead to substantially higher cost, whereas the impact of $W$ becomes negligible for larger $N$.}
    \label{tab:runtime}
    \setlength\tabcolsep{1.45pt}
    \renewcommand{\arraystretch}{0.8}
    \begin{tabular}{lcccccccc}
        \toprule

        & \multicolumn{4}{c}{1024px~$\times$~640px}
        & \multicolumn{4}{c}{640px~$\times$~280px} \\
        \cmidrule(lr){2-5}
        \cmidrule(lr){6-9}
        & $W=10$ & $W=20$ & $W=30$ & $W=40$ & $W=10$ & $W=20$ & $W=30$ & $W=40$ \\

        \midrule

        $N=3$  & 1.28s & 1.75s & 2.34s & 2.77s
           & 0.50s & 0.66s & 0.80s & 0.99s \\

        $N=5$  & 0.50s & 0.55s & 0.59s & 0.63s
           & 0.27s & 0.28s & 0.29s & 0.30s \\

        $N=7$  & -- & 0.43s & 0.46s & 0.48s
           & -- & 0.24s & 0.25s & 0.26s \\

        $N=9$  & -- & 0.41s & 0.42s & 0.43s
           & -- & 0.20s & 0.21s & 0.22s \\

        $N=11$ & -- & -- & 0.41s & 0.42s
           & -- & -- & 0.19s & 0.19s \\

        $N=13$ & -- & -- & 0.40s & 0.41s
           & -- & -- & 0.19s & 0.20s \\

        \bottomrule
    \end{tabular}
\end{table}

The two parameters also jointly affect computational efficiency. A larger $W$ provides access to a broader set of candidate patches and may improve reconstruction quality, but it also enlarges the contextual search space. The number of candidate patches within the search band scales approximately as $\CMcal{O}(W/N)$: increasing $W$ expands the search space, whereas increasing $N$ reduces the number of patches but increases the dimensionality of each patch comparison. 
As shown in \tabref{tab:runtime}, the runtime is primarily determined by $N$. At a resolution of $1024\times640$, increasing $N$ from $3$ to $5$ reduces the runtime from $1.75$s to $0.55$s when $W=20$, while the corresponding reduction at $640\times280$ is from $0.66$s to $0.28$s. By contrast, the influence of $W$ becomes progressively weaker as $N$ increases. For example, at $1024\times640$, increasing $W$ from $20$ to $40$ raises the runtime from $1.75$s to $2.77$s for $N=3$, but only from $0.43$s to $0.48$s for $N=7$ and from $0.41$s to $0.43$s for $N=9$. Similar trends are observed at the lower resolution. These results indicate that when $N$ is small, the increased number of valid patches and patch comparisons has a substantially greater impact on runtime than the increased dimensionality of individual patch comparisons.

Considering the trade-off among contextual richness, target coverage, search range, and efficiency, we set $N=7$ and $W=20$. This configuration achieves an average target coverage ratio of 87.53\%, provides a sufficiently broad contextual search range, and maintains an acceptable runtime across typical image resolutions.

\begin{figure*}[t!]
    \centering
    \includegraphics[width=1\linewidth]{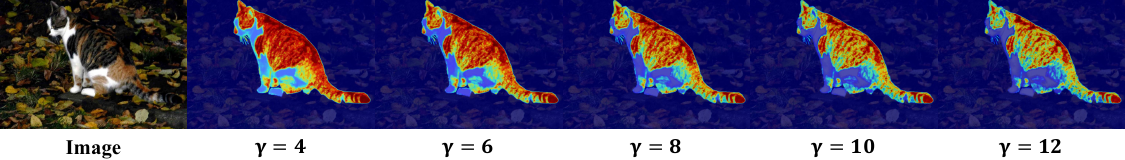}
    \caption{\textbf{Effect of $\gamma$}.
    $\gamma$ controls the curvature of the mapping: larger values suppress the overall affinity estimates, making the distribution more conservative, whereas smaller values produce a more spread out distribution, but may also lead to affinity overestimation.
    }
    \label{fig:gamma}
    \vspace{-9pt}
\end{figure*}

\begin{figure*}[t!]
    \centering
    \includegraphics[width=1\linewidth]{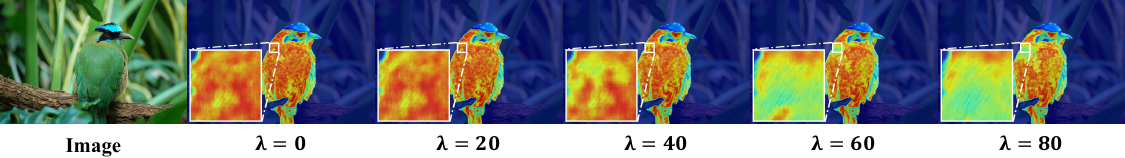}
    \caption{\textbf{Effect of $\lambda$}.
    $\lambda$ controls the balance between color similarity and spatial proximity during patch matching: larger values enforce stronger spatial constraints, leading to larger reconstruction discrepancies and thus lower affinity estimates for local regions.
    }
    \label{fig:lambda}
\end{figure*}

\begin{table*}[t!]
    \centering
    \caption{
    \textbf{Effectiveness of contextual affinity 
    $\mD$.} Removing $\mD$ generally degrades 
    performance across the evaluated meta-measures. 
    MM: Meta-Measure; 
    $^{*}$0.01\%~~: $\le$0.01\%.
    The best and second-best results are highlighted in \red{red} and \blue{blue}, respectively.
    }
    \setlength\tabcolsep{4.2pt}
    \renewcommand\arraystretch{0.8}
    \begin{tabular}{cccccccccccccc}
    \toprule
    \multirow[c]{2}{*}[-1.0ex]{Metric} & \multicolumn{1}{c}{MM\#1} 
    & \multicolumn{3}{c}{MM\#2} 
    & \multicolumn{3}{c}{MM\#3} 
    & \multicolumn{3}{c}{MM\#4: Erode} 
    & \multicolumn{3}{c}{MM\#4: Dilate}\\
    \cmidrule(lr){2-2} \cmidrule(lr){3-5} \cmidrule(lr){6-8} 
    \cmidrule(lr){9-11} \cmidrule(lr){12-14}
    & CamoHR 
    & COD10K & NC4K & Trans10K 
    & COD10K & NC4K & Trans10K 
    & COD10K & NC4K & Trans10K 
    & COD10K & NC4K & Trans10K \\ \midrule
    IoU  & 7.93\%  & 0.05\%  & 0.06\%  & 0.08\%  & 0.55\%  & 0.39\%  & 3.00\%  & 2.76\%  & 1.99\% & 0.48\%  & 1.67\%  & 1.32\%   & 0.46\% \\
    $F_\beta$  & 17.43\%~~ & 0.34\%  & 0.32\%  & 0.14\%  & 1.48\%  & 2.00\%  & 7.65\%   & 3.28\%   & 2.54\%  & 0.46\%  & 2.31\%   & 1.46\%  & 0.40\%  \\
    $F_{\beta}^{\omega}$  & 5.50\%  & 0.09\%                         & 0.13\%                         & 0.05\%                         & \multicolumn{1}{l}{\red{$^{*}$0.01\%~~}} & \multicolumn{1}{l}{\red{$^{*}$0.01\%~~}} & \multicolumn{1}{l}{\red{~$^{*}$0.01\%}}                        & 2.12\%                         & 1.49\%                         & 0.44\%                         & 1.13\%                    & 0.83\%                    & 0.24\%                    \\
    $S_\alpha$                                & 8.25\%     & 0.09\%                         & 0.06\%                         & 0.54\%                         & 10.47\%~~                        & 8.43\%                         & 0.53\%                        & 1.34\%                         & 0.97\%                         & 0.31\%                         & 0.85\%                    & 0.77\%                    & 0.20\%                    \\
    $E_\phi$                                  & 18.00\%~~     & 3.46\%                         & 2.37\%                         & 1.62\%                         & 13.73\%~~                        & 15.64\%~~                        & 5.62\%                         & 1.79\%                         & 1.16\%                         & 0.38\%                         & 1.16\%                    & 0.93\%                    & 0.18\%                    \\ \midrule
    \rowcolor[HTML]{EFEFEF}
    $C_{\beta}$
    & \blue{4.00\%}
    & \multicolumn{1}{l}{\red{$^{*}$0.01\%~~}} & \blue{0.06\%} & \blue{0.03\%}
    & \multicolumn{1}{l}{\red{$^{*}$0.01\%~~}} & \multicolumn{1}{l}{\red{$^{*}$0.01\%~~}} & \multicolumn{1}{l}{\red{~$^{*}$0.01\%}}
    & \blue{1.33\%} & \blue{0.88\%} & \blue{0.30\%} 
    & \blue{0.84\%} & \red{0.61\%} & \blue{0.15\%} \\
    \rowcolor[HTML]{EFEFEF}
    $C_{\beta}^{\omega}$ 
    & \red{3.25\%} 
    & \multicolumn{1}{l}{\red{$^{*}$0.01\%~~}} & \red{0.03\%} & \red{0.03\%} 
    & \multicolumn{1}{l}{\red{$^{*}$0.01\%~~}} & \multicolumn{1}{l}{\red{$^{*}$0.01\%~~}} & \multicolumn{1}{l}{\red{~$^{*}$0.01\%}}
    & \red{1.21\%} & \red{0.80\%} & \red{0.29\%} 
    & \red{0.80\%} & \blue{0.64\%} & \red{0.14\%} \\
    \bottomrule
    \end{tabular}
    \label{tab:ablation}
\end{table*}
    
\subsubsection{Nonlinearity Coefficient \texorpdfstring{$\gamma$}{gamma} and Spatial Weight \texorpdfstring{$\lambda$}{lambda}}\label{sec:gamma_lambda}

The nonlinearity coefficient $\gamma$ controls the curvature of the mapping from color difference to contextual affinity: larger values suppress the overall affinity estimates, making the distribution more conservative, while smaller values produce a more spread-out distribution. The spatial weighting factor $\lambda$ controls the balance between color similarity and spatial proximity during patch matching: larger values enforce stronger spatial constraints, leading to larger reconstruction discrepancies and thus lower affinity estimates in local regions.

Since both hyperparameters govern the distribution of $\mD$ rather than a directly optimizable objective, we select their values by examining the resulting affinity maps and verifying their alignment with human intuition about camouflage difficulty. As visualized in \figref{fig:gamma} and \figref{fig:lambda}, we vary $\gamma \in \{4,6,8,10,12\}$ and $\lambda \in \{0, 20, 40, 60, 80\}$ independently and inspect the corresponding $\mD$ distributions. We select $\gamma=8$ and $\lambda=20$ as they consistently produce affinity maps in which highly concealed regions receive higher values and more visually distinguishable regions receive lower values, in agreement with human perception.

\subsubsection{Scale Factor \texorpdfstring{$\beta$}{beta}}\label{sec:beta}

The scale factor $\beta$ balances the contributions of the forward term $F_m$ and the reverse term $R_{\omega}$. Since an evaluation metric should align with human perceptual judgment, we select $\beta$ based on the agreement between metric-induced and human-annotated rankings. Specifically, we evaluate $C_{\beta}^{\omega}$ on the CamoHR validation set with $\beta^2 \in \{0.6, 0.8, 1.0, 1.2, 1.4\}$. Following the protocol in \secref{sec:mm1}, we measure ranking disagreement using $\theta = 1-\rho$, where $\rho$ denotes the Spearman rank correlation coefficient and a lower $\theta$ indicates stronger agreement with human judgment. 
$\beta^2=1.2$ achieves the best consistency across all evaluated parameter settings and is adopted as the default configuration.

\subsection{Contextual Affinity}\label{sec:ablation_affinity}

To validate the contribution of the contextual affinity $\mD$, we compare the Context-measure $C_{\beta}^{\omega}$ against a degraded variant in which $\mD$ is removed, \ie, $\mD=\mathbf{0}$. Under this setting, the weighted reverse term $R_{\omega}$ degrades to the unweighted $R_m$, and the metric reduces to a context-unaware loop that treats all GT pixels equally regardless of their concealment difficulty. We denote this variant as $C_{\beta}$, whose scale factor $\beta$ is independently selected on the CamoHR validation set following the same protocol as \secref{sec:beta}, yielding $\beta^2=1.0$ as the optimal configuration.

As shown in \tabref{tab:ablation}, incorporating contextual affinity improves performance across the four meta-measures on all three datasets, with only a few exceptions.
The most pronounced improvement is observed in MM\#1: on the CamoHR test set, $C_{\beta}$ achieves a ranking disagreement $\theta$ of 4.00\%, whereas $C_{\beta}^{\omega}$ further reduces it to 3.25\%, demonstrating that incorporating contextual affinity leads to rankings more consistent with human perception. The consistent gains in MM\#2, MM\#3, and MM\#4 further demonstrate that $\mD$ improves evaluation reliability.

\subsection{Pixel Correlation Framework}\label{sec:ablation_framework}

To examine the effectiveness of the proposed pixel correlation framework, we also consider $C_{\beta}$, which retains only the perception cycle built upon the pixel correlation framework. As shown in \tabref{tab:ablation}, $C_{\beta}$ already outperforms all existing metrics across most meta-measures in camouflaged scenarios. 
This result demonstrates that the proposed pixel correlation framework alone provides a stronger foundation for segmentation evaluation.

\begin{figure}[t!]
    \centering
    \includegraphics[width=1\linewidth]{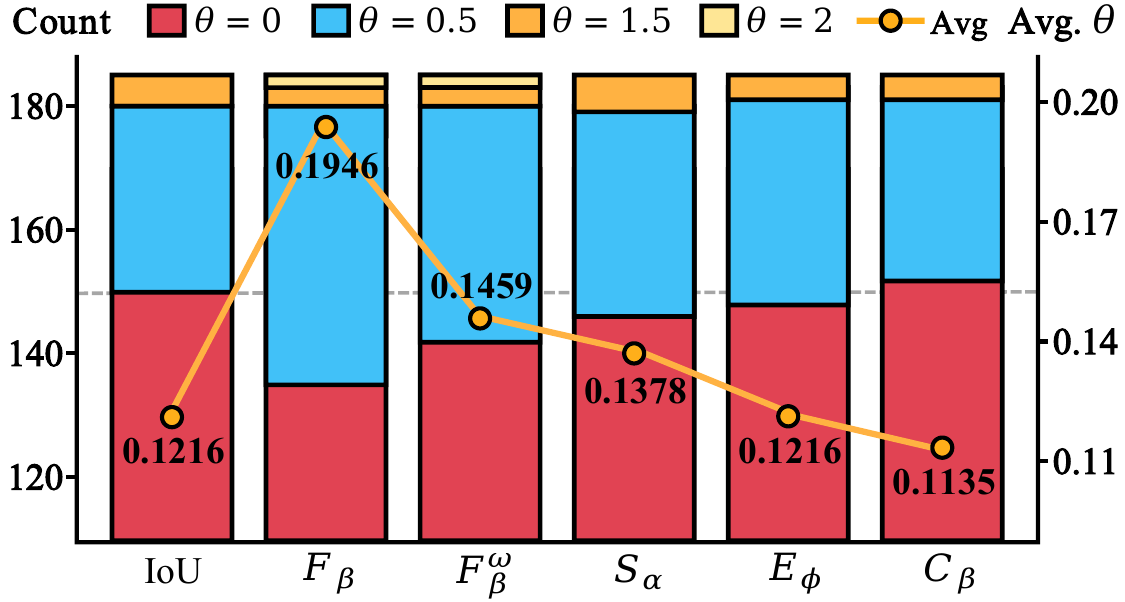}
    \caption{
    \textbf{Effect of pixel correlation framework.}
    }
    \label{fig:fmdatabase_data}
\end{figure}

\begin{figure}[t!]
    \centering
    \includegraphics[width=0.9\linewidth]{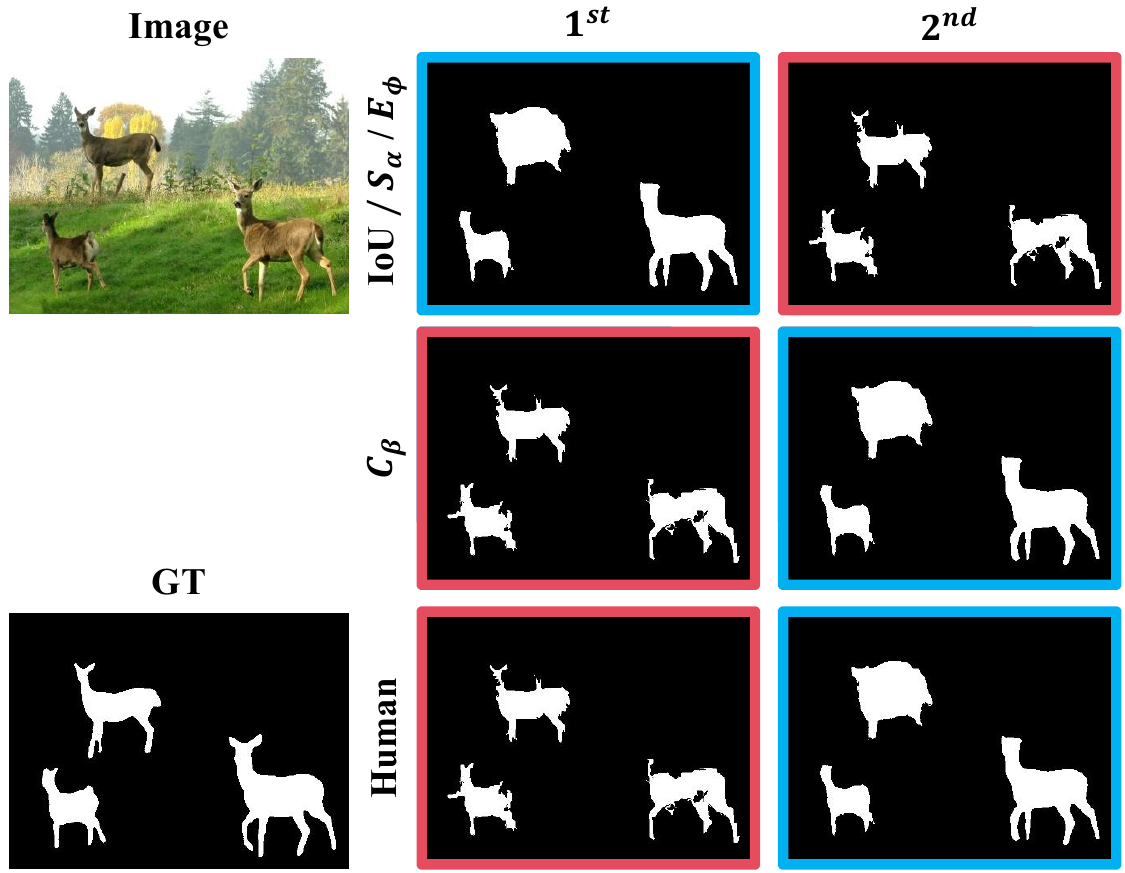}
    \caption{
    \textbf{Qualitative comparison in salient scenarios.} Two predicted FMs are ranked by human judgment: red-bordered first, blue-bordered second. 
    Two widely used metrics ($S_{\alpha}$ and $E_{\phi}$) fail to reproduce this order, whereas our $C_{\beta}$ correctly preserves it.
    }
    \label{fig:saliency}
\end{figure}

Because the pixel correlation framework is formulated around the fundamental relationship between a predicted FM and its GT, it provides a general foundation for segmentation evaluation. To further examine this generality beyond camouflaged scenarios, we evaluate $C_{\beta}$ on salient object segmentation~\cite{cheng2014global}, a closely related object segmentation task in which many metrics used in COS were originally developed or validated.
Following the experimental protocol described in \secref{sec:mm1}, we assess the agreement between metric-induced and human-annotated rankings on FMDatabase~\cite{fan2018enhanced}, which contains binary foreground maps from salient scenarios together with human quality rankings. As shown in \figref{fig:fmdatabase_data}, $C_{\beta}$ yields the $\theta$ of 11.35\%, outperforming all existing metrics, including $S_{\alpha}$ (13.78\%) and $E_{\phi}$ (12.16\%). 
The qualitative comparisons in \figref{fig:saliency} further illustrate its ability to distinguish predicted FM quality across salient scenarios, demonstrating the broader applicability of the pixel correlation framework.

\begin{table}[t!]
    \centering
    \caption{\textbf{Effect of color space on contextual affinity.}}
    \setlength\tabcolsep{6pt}
    \renewcommand\arraystretch{0.8}
    \begin{tabular}{ccccccc}
        \toprule
        CamoHR & LAB & RGB & HSV & HSL & YUV & YCbCr\\ \midrule
        Validation & 7.00\% & 7.00\% & 7.00\% & 7.00\% & 8.00\% & 8.00\% \\
        Test & 3.25\% & 4.00\% & 3.75\% & 3.75\% & 4.00\% & 3.25\% \\ \bottomrule
    \end{tabular}
    \label{tab:colorspace}
\end{table}

\subsection{Color Space}\label{sec:ablation_colorspace}

As reported in \tabref{tab:colorspace}, to investigate the impact of color representation on contextual affinity estimation, we evaluate six commonly used color spaces, namely RGB, HSV, HSL, YUV, YCbCr, and LAB. 
Following the experimental protocol described in \secref{sec:mm1}, we conduct experiments on the CamoHR dataset and report the ranking disagreement $\theta$ on both its validation and test sets.
For each variant, the corresponding color space is consistently used for both patch matching in Stage I and color-difference computation in Stage II, while all other settings remain unchanged to ensure a fair comparison.
Although the alternative color spaces characterize color information from different perspectives, their numerical distances are generally less consistent with human perception of color differences.
In contrast, LAB provides better perceptual correspondence between numerical color differences and human color perception, enabling a more reliable estimation of local target--context affinity. Thus, it produces more reliable contextual affinity $\mD$ and achieves stronger agreement with human judgment.

\begin{figure}[t!]
    \centering
    \includegraphics[width=1\linewidth]{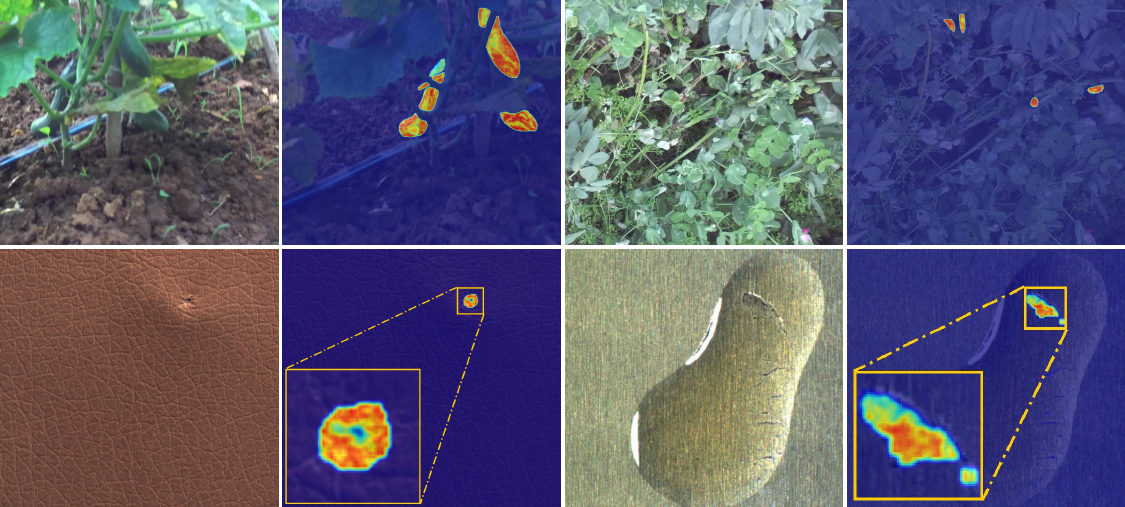}
    \caption{
    \textbf{Generalization to more applications.} We present qualitative examples of contextual affinity for two camouflage-related application scenarios.
    The first row presents crop detection, while the second row presents surface defect detection.
    }
    \label{fig:scenarios}
\end{figure}

\section{Generalization to More Applications}
Beyond polyp and mirror segmentation, Context-measure has the potential to support a broader range of camouflage-related applications. As shown in \figref{fig:scenarios}, our quantification method produces meaningful pixel-wise contextual affinity maps for both crop detection and surface defect detection without modifying any hyperparameters.
In these scenarios, targets may exhibit weak boundaries, subtle appearance differences, or strong visual similarity to their surroundings, making conventional context-blind evaluation insufficient to fully characterize segmentation difficulty.
Similar challenges arise in wildlife monitoring~\cite{fan2020camouflaged}, medical lesion segmentation~\cite{jafari2016skin}, underwater target detection~\cite{wang2026expose}, and transparent or reflective object segmentation~\cite{kalra2020deep}.
Although the specific notion of camouflage may vary across applications, these tasks commonly require visually inconspicuous targets to be distinguished from their surroundings. 
The successful affinity estimation suggests that our contextual quantification method can capture such task-dependent visual ambiguity beyond conventional COS scenarios. By further adapting the spatial quantification process and the use of contextual affinity to the characteristics and evaluation objectives of each task, Context-measure may offer broad applicability across diverse camouflage-related domains.

\section{Conclusion}\label{sec:conclusion}  

In this paper, we identify two major drawbacks of existing context-blind metrics for camouflaged object segmentation: the \textit{Dimension Flaw}, which prevents them from differentiating GT pixels by contextual difficulty, and the \textit{Range Flaw}, which prevents them from capturing full-range pixel dependencies. To address these drawbacks, we propose \textbf{Context-measure} ($C_{\beta}^{\omega}$), the first COS-specific evaluation metric. Context-measure augments the GT with pixel-wise contextual affinity $\mD$ and evaluates the predicted FM through a perception cycle constructed upon a probabilistic pixel correlation framework. We further curate \textbf{CamoHR}, a human-ranked dataset containing 750 predicted FMs, to benchmark metric consistency with human perception. Extensive experiments across four meta-measures demonstrate that Context-measure comprehensively outperforms widely adopted metrics, achieving a 41\% relative improvement in consistency with human judgment.

In summary, an ideal evaluation metric should not only assess model performance but also reveal existing model drawbacks and guide future methodological development. We hope that Context-measure can inspire new perspectives on segmentation evaluation and promote further progress in camouflage-oriented research.

\ifCLASSOPTIONcaptionsoff
  \newpage
\fi

{
\bibliographystyle{IEEEtran}
\bibliography{bibliography}

@String(AAAI = {AAAI})

@String(CVPR  = {CVPR})

@String(ICCV  = {ICCV})

@String(ICIP  = {ICIP})

@String(ECCV  = {ECCV})

@String(ICPR  = {ICPR})

@String(ICLR  = {ICLR})

@String(ICML  = {ICML})

@String(IJCAI = {IJCAI})

@String(MMM = {MMM})

@String(ACMMM = {ACM MM})

@String(MICCAI= {MICCAI})

@String(CRV= {CRV})

@String(MIR = {MIR})

@String(CARS = {CARS})

@String(TPAMI = {IEEE TPAMI})

@String(TIP = {IEEE TIP})

@String(IJCV  = {IJCV})

@String(TCSVT = {IEEE TCSVT})

@String(PR = {PR})

@String(JAIR = {CAAI AIR})

@String(AS = {ApplSci})

@String(CMIG = {CMIG})

@String(CVMJ = {CVMJ})

@String(JI = {J. Imaging})

@String(TMI = {IEEE TMI})

@String(JHE = {JHE})

@String(SSI = {SSI})

@String(TMM = {IEEE TMM})

@String(VI = {VI})

@String(CVIU = {CVIU})

@article{alpert2011image,
  title={Image segmentation by probabilistic bottom-up aggregation and cue integration},
  author={Alpert, Sharon and Galun, Meirav and Brandt, Achi and Basri, Ronen},
  journal=TPAMI,
  volume={34},
  number={2},
  pages={315--327},
  year={2011},
}

@inproceedings{margolin2014evaluate,
  title={How to evaluate foreground maps?},
  author={Margolin, Ran and Zelnik-Manor, Lihi and Tal, Ayellet},
  booktitle=CVPR,
  //pages={248--255},
  year={2014}
}

@inproceedings{fan2017structure,
  title={Structure-measure: A new way to evaluate foreground maps},
  author={Fan, Deng-Ping and Cheng, Ming-Ming and Liu, Yun and Li, Tao and Borji, Ali},
  booktitle=ICCV,
  //pages={4548--4557},
  year={2017}
}

@article{cheng2014global,
  title={Global contrast based salient region detection},
  author={Cheng, Ming-Ming and Mitra, Niloy J and Huang, Xiaolei and Torr, Philip HS and Hu, Shi-Min},
  journal=TPAMI,
  volume={37},
  number={3},
  pages={569--582},
  year={2014},
  publisher={Ieee}
}

@inproceedings{fan2018enhanced,
  title={Enhanced-alignment measure for binary foreground map evaluation},
  author={Fan, Deng-Ping and Gong, Cheng and Cao, Yang and Ren, Bo and Cheng, Ming-Ming and Borji, Ali},
  booktitle=IJCAI,
  year={2018},
  //pages={698--704}
}

@article{arbelaez2010contour,
  title={Contour detection and hierarchical image segmentation},
  author={Arbelaez, Pablo and Maire, Michael and Fowlkes, Charless and Malik, Jitendra},
  journal=TPAMI,
  volume={33},
  number={5},
  pages={898--916},
  year={2010},
}

@article{best1975algorithm,
  title={Algorithm AS 89: the upper tail probabilities of Spearman's rho},
  author={Best, DJ and Roberts, DE},
  journal={Journal of the Royal Statistical Society. Series C (Applied Statistics)},
  volume={24},
  number={3},
  pages={377--379},
  year={1975},
  publisher={JSTOR}
}

@inproceedings{lamdouar2023making,
  title={The making and breaking of camouflage},
  author={Lamdouar, Hala and Xie, Weidi and Zisserman, Andrew},
  booktitle=ICCV,
  //pages={832--842},
  year={2023}
}

@inproceedings{fan2020camouflaged,
  title={Camouflaged object detection},
  author={Fan, Deng-Ping and Ji, Ge-Peng and Sun, Guolei and Cheng, Ming-Ming and Shen, Jianbing and Shao, Ling},
  booktitle=CVPR,
  //pages={2777--2787},
  year={2020}
}

@inproceedings{he2023camouflaged,
  title={Camouflaged object detection with feature decomposition and edge reconstruction},
  author={He, Chunming and Li, Kai and Zhang, Yachao and Tang, Longxiang and Zhang, Yulun and Guo, Zhenhua and Li, Xiu},
  booktitle=CVPR,
  //pages={22046--22055},
  year={2023}
}

@inproceedings{huang2023feature,
  title={Feature shrinkage pyramid for camouflaged object detection with transformers},
  author={Huang, Zhou and Dai, Hang and Xiang, Tian-Zhu and Wang, Shuo and Chen, Huai-Xin and Qin, Jie and Xiong, Huan},
  booktitle=CVPR,
  //pages={5557--5566},
  year={2023}
}

@inproceedings{hu2023high,
  title={High-resolution iterative feedback network for camouflaged object detection},
  author={Hu, Xiaobin and Wang, Shuo and Qin, Xuebin and Dai, Hang and Ren, Wenqi and Luo, Donghao and Tai, Ying and Shao, Ling},
  booktitle=AAAI,
  //pages={881--889},
  year={2023}
}

@inproceedings{zha2026seeing,
  title={Seeing beyond illusion: Generalized and efficient mirror detection},
  author={Zha, Mingfeng and Wang, Guoqing and Li, Tianyu and Dong, Wei and Wang, Peng and Yang, Yang},
  booktitle=AAAI,
  //volume={40},
  //number={15},
  //pages={12331--12339},
  year={2026}
}

@inproceedings{ravi2024sam2,
  title={SAM 2: Segment Anything in Images and Videos},
  author={Ravi, Nikhila and Gabeur, Valentin and Hu, Yuan-Ting and Hu, Ronghang and Ryali, Chaitanya and Ma, Tengyu and Khedr, Haitham and R{\"a}dle, Roman and Rolland, Chloe and Gustafson, Laura and Mintun, Eric and Pan, Junting and Alwala, Kalyan Vasudev and Carion, Nicolas and Wu, Chao-Yuan and Girshick, Ross and Doll{\'a}r, Piotr and Feichtenhofer, Christoph},
  booktitle=ICLR,
  year={2025}
}

@article{fan2021concealed,
  title={Concealed object detection},
  author={Fan, Deng-Ping and Ji, Ge-Peng and Cheng, Ming-Ming and Shao, Ling},
  journal=TPAMI,
  volume={44},
  number={10},
  pages={6024--6042},
  year={2021},
  publisher={IEEE}
}

@inproceedings{pang2022zoom,
  title={Zoom in and out: A mixed-scale triplet network for camouflaged object detection},
  author={Pang, Youwei and Zhao, Xiaoqi and Xiang, Tian-Zhu and Zhang, Lihe and Lu, Huchuan},
  booktitle=CVPR,
  //pages={2160--2170},
  year={2022}
}

@inproceedings{lv2021simultaneously,
  title={Simultaneously localize, segment and rank the camouflaged objects},
  author={Lv, Yunqiu and Zhang, Jing and Dai, Yuchao and Li, Aixuan and Liu, Bowen and Barnes, Nick and Fan, Deng-Ping},
  booktitle=CVPR,
  //pages={11591--11601},
  year={2021}
}

@inproceedings{sun2022boundary,
  title={Boundary-Guided Camouflaged Object Detection},
  author={Sun, Yujia and Wang, Shuo and Chen, Chenglizhao and Xiang, Tian Zhu},
  booktitle=IJCAI,
  //pages={1335--1341},
  year={2022},
  //organization={International Joint Conferences on Artificial Intelligence}
}

@inproceedings{xie2020segmenting,
  title={Segmenting transparent objects in the wild},
  author={Xie, Enze and Wang, Wenjia and Wang, Wenhai and Ding, Mingyu and Shen, Chunhua and Luo, Ping},
  booktitle=ECCV,
  //pages={696--711},
  year={2020},
}

@article{everingham2010pascal,
  title={The pascal visual object classes (voc) challenge},
  author={Everingham, Mark and Van Gool, Luc and Williams, Christopher KI and Winn, John and Zisserman, Andrew},
  journal=IJCV,
  volume={88},
  number={2},
  pages={303--338},
  year={2010},
  publisher={Springer}
}

@article{fan2021structure,
  title={Structure-measure: A new way to evaluate foreground maps},
  author={Cheng, Ming-Ming and Fan, Deng-Ping},
  journal=IJCV,
  volume={129},
  number={9},
  pages={2622--2638},
  year={2021},
  publisher={Springer}
}

@article{fan2021cognitive,
  title={Cognitive vision inspired object segmentation metric and loss function},
  author={Fan, Deng-Ping and Ji, Ge-Peng and Qin, Xuebin and Cheng, Ming-Ming},
  journal=SSI,
  volume={6},
  number={6},
  pages={5},
  year={2021},
  publisher={Science China Press}
}

@article{ji2023deep,
  title={Deep gradient learning for efficient camouflaged object detection},
  author={Ji, Ge-Peng and Fan, Deng-Ping and Chou, Yu-Cheng and Dai, Dengxin and Liniger, Alexander and Van Gool, Luc},
  journal=MIR,
  volume={20},
  number={1},
  pages={92--108},
  year={2023},
  publisher={Springer}
}

@inproceedings{wang2024depth,
  title={Depth-Aware Concealed Crop Detection in Dense Agricultural Scenes},
  author={Wang, Liqiong and Yang, Jinyu and Zhang, Yanfu and Wang, Fangyi and Zheng, Feng},
  booktitle=CVPR,
  //pages={17201--17211},
  year={2024}
}

@article{zha2024dual,
  title={Dual domain perception and progressive refinement for mirror detection},
  author={Zha, Mingfeng and Fu, Feiyang and Pei, Yunqiang and Wang, Guoqing and Li, Tianyu and Tang, Xiongxin and Yang, Yang and Shen, Heng Tao},
  journal=TCSVT,
  volume={34},
  number={11},
  pages={11942--11953},
  year={2024},
  publisher={IEEE}
}

@article{zhou2024decoupling,
  title={Decoupling and integration network for camouflaged object detection},
  author={Zhou, Xiaofei and Wu, Zhicong and Cong, Runmin},
  journal=TMM,
  volume={26},
  pages={7114--7129},
  year={2024},
  publisher={IEEE}
}

@article{yin2024camoformer,
  title={Camoformer: Masked separable attention for camouflaged object detection},
  author={Yin, Bowen and Zhang, Xuying and Fan, Deng-Ping and Jiao, Shaohui and Cheng, Ming-Ming and Van Gool, Luc and Hou, Qibin},
  journal=TPAMI,
  volume={46},
  number={12},
  pages={10362--10374},
  year={2024},
  publisher={IEEE}
}

@article{pang2024zoomnext,
  title={Zoomnext: A unified collaborative pyramid network for camouflaged object detection},
  author={Pang, Youwei and Zhao, Xiaoqi and Xiang, Tian-Zhu and Zhang, Lihe and Lu, Huchuan},
  journal=TPAMI,
  volume={46},
  number={12},
  pages={9205--9220},
  year={2024},
  publisher={IEEE}
}

@inproceedings{ye2025escnet,
  title={Escnet: Edge-semantic collaborative network for camouflaged object detection},
  author={Ye, Sheng and Chen, Xin and Zhang, Yan and Lin, Xianming and Cao, Liujuan},
  booktitle=ICCV,
  //pages={20053--20063},
  year={2025}
}

@inproceedings{he2025run,
  title={RUN: Reversible Unfolding Network for Concealed Object Segmentation},
  author={He, Chunming and Zhang, Rihan and Xiao, Fengyang and Fang, Chengyu and Tang, Longxiang and Zhang, Yulun and Kong, Linghe and Fan, Deng-Ping and Li, Kai and Farsiu, Sina},
  booktitle=ICML,
  year={2025}
}

@inproceedings{yu2025sam,
  title={Sam-ttt: Segment anything model via reverse parameter configuration and test-time training for camouflaged object detection},
  author={Yu, Zhenni and Zhao, Li and Xiao, Guobao and Zhang, Xiaoqin},
  booktitle=ACMMM,
  //pages={4030--4038},
  year={2025}
}

@article{song2025continuous,
  title={Continuous Feature Representation for Camouflaged Object Detection},
  author={Song, Ze and Kang, Xudong and Wei, Xiaohui and Liu, Jinyang and Lin, Zheng and Li, Shutao},
  journal=TIP,
  volume={34},
  number={},
  pages={5672--5685},
  year={2025}
}

@article{hao2025simple,
  title={A Simple Yet Effective Network Based on Vision Transformer for Camouflaged Object and Salient Object Detection},
  author={Hao, Chao and Yu, Zitong and Liu, Xin and Xu, Jun and Yue, Huanjing and Yang, Jingyu},
  journal=TIP,
  volume={34},
  number={},
  pages={608--622},
  year={2025},
  publisher={IEEE}
}

@article{li2026learning,
  title={Learning Compact Representations With an Information Bottleneck for Camouflaged Object Detection},
  author={Li, Guanyi and Zhang, Junjie and Gao, Rui and Yuan, Wubang and Jin, Gloria and Zeng, Dan},
  journal=TMM,
  volume={28},
  number={},
  pages={360--372},
  year={2026}
}

@article{zhao2025spatial,
  title={Spatial-Frequency Collaborative Learning for Camouflaged Object Detection},
  author={Zhao, Rui and Wang, Mengyin and Wang, Fasheng and Sun, Fuming and Li, Haojie},
  journal=TMM,
  volume={27},
  number={},
  pages={7756--7768},
  year={2025}
}

@article{sun2025conditional,
  title={Conditional diffusion models for camouflaged and salient object detection},
  author={Sun, Ke and Chen, Zhongxi and Lin, Xianming and Sun, Xiaoshuai and Liu, Hong and Ji, Rongrong},
  journal=TPAMI,
  volume={47},
  number={4},
  pages={2833--2848},
  year={2025},
  publisher={IEEE}
}

@article{wang2026gbnet,
  title={GBNet: Gated Boundary-Aware Network for Camouflaged Object Detection.},
  author={Wang, X and Yao, F and Zhong, G and Cai, Q and Wang, S and Kwok, JT},
  journal=TIP,
  volume={35},
  number={},
  pages={5297--5310},
  year={2026}
}

@article{chen2026learn,
  title={Learn From Examples: In-Context Learning for Camouflaged Object Detection},
  author={Chen, Chunyuan and Liang, Weiyun and Du, Ji and Xu, Jing and Li, Ping and Wang, Guiling},
  journal=TIP,
  volume={35},
  number={},
  pages={3793--3806},
  year={2026},
  publisher={IEEE}
}

@article{jaccard1901etude,
  title={{\'E}tude comparative de la distribution florale dans une portion des Alpes et des Jura},
  author={Jaccard, Paul},
  journal={Bull Soc Vaudoise Sci Nat},
  volume={37},
  pages={547--579},
  year={1901}
}

@inproceedings{ahmadzadeh2021multiscale,
  title={Multiscale iou: A metric for evaluation of salient object detection with fine structures},
  author={Ahmadzadeh, Azim and Kempton, Dustin J and Chen, Yang and Angryk, Rafal A},
  booktitle=ICIP,
  //pages={684--688},
  year={2021},
}

@inproceedings{fan2020pranet,
  title={Pranet: Parallel reverse attention network for polyp segmentation},
  author={Fan, Deng-Ping and Ji, Ge-Peng and Zhou, Tao and Chen, Geng and Fu, Huazhu and Shen, Jianbing and Shao, Ling},
  booktitle=MICCAI,
  //pages={263--273},
  year={2020},
  //organization={Springer}
}

@inproceedings{li2021uncertainty,
  title={Uncertainty-aware joint salient object and camouflaged object detection},
  author={Li, Aixuan and Zhang, Jing and Lv, Yunqiu and Liu, Bowen and Zhang, Tong and Dai, Yuchao},
  booktitle=CVPR,
  //pages={10071--10081},
  year={2021}
}

@inproceedings{mei2021camouflaged,
  title={Camouflaged object segmentation with distraction mining},
  author={Mei, Haiyang and Ji, Ge-Peng and Wei, Ziqi and Yang, Xin and Wei, Xiaopeng and Fan, Deng-Ping},
  booktitle=CVPR,
  //pages={8772--8781},
  year={2021}
}

@inproceedings{yang2021uncertainty,
  title={Uncertainty-guided transformer reasoning for camouflaged object detection},
  author={Yang, Fan and Zhai, Qiang and Li, Xin and Huang, Rui and Luo, Ao and Cheng, Hong and Fan, Deng-Ping},
  booktitle=ICCV,
  //pages={4146--4155},
  year={2021}
}

@inproceedings{sun2021context,
  title={Context-aware Cross-level Fusion Network for Camouflaged Object Detection},
  author={Sun, Yujia and Chen, Geng and Zhou, Tao and Zhang, Yi and Liu, Nian},
  booktitle=IJCAI,
  //pages={1025--1031},
  year={2021},
  //organization={International Joint Conferences on Artificial Intelligence}
}

@inproceedings{zhu2021inferring,
  title={Inferring camouflaged objects by texture-aware interactive guidance network},
  author={Zhu, Jinchao and Zhang, Xiaoyu and Zhang, Shuo and Liu, Junnan},
  booktitle=AAAI,
  //volume={35},
  //number={4},
  //pages={3599--3607},
  year={2021}
}

@inproceedings{lin2014microsoft,
  title={Microsoft coco: Common objects in context},
  author={Lin, Tsung-Yi and Maire, Michael and Belongie, Serge and Hays, James and Perona, Pietro and Ramanan, Deva and Doll{\'a}r, Piotr and Zitnick, C Lawrence},
  booktitle=ECCV,
  //pages={740--755},
  year={2014},
  //organization={Springer}
}

@inproceedings{milletari2016v,
  title={V-net: Fully convolutional neural networks for volumetric medical image segmentation},
  author={Milletari, Fausto and Navab, Nassir and Ahmadi, Seyed-Ahmad},
  booktitle={3DV},
  //pages={565--571},
  year={2016},
}

@inproceedings{zhao2019optimizing,
  title={Optimizing the F-measure for threshold-free salient object detection},
  author={Zhao, Kai and Gao, Shanghua and Wang, Wenguan and Cheng, Ming-Ming},
  booktitle=ICCV,
  //pages={8849--8857},
  year={2019}
}

@inproceedings{sudre2017generalised,
  title={Generalised dice overlap as a deep learning loss function for highly unbalanced segmentations},
  author={Sudre, Carole H and Li, Wenqi and Vercauteren, Tom and Ourselin, Sebastien and Jorge Cardoso, M},
  booktitle={DLMIA-w},
  //pages={240--248},
  year={2017},
  //organization={Springer}
}

@article{xiong2026sam2,
  title={Sam2-unet: Segment anything 2 makes strong encoder for natural and medical image segmentation},
  author={Xiong, Xinyu and Wu, Zihuang and Tan, Shuangyi and Li, Wenxue and Tang, Feilong and Chen, Ying and Li, Siying and Ma, Jie and Li, Guanbin},
  journal=VI,
  volume={4},
  number={1},
  pages={2},
  year={2026},
  publisher={Springer}
}

@inproceedings{ronneberger2015u,
  title={U-net: Convolutional networks for biomedical image segmentation},
  author={Ronneberger, Olaf and Fischer, Philipp and Brox, Thomas},
  booktitle=MICCAI,
  //pages={234--241},
  year={2015},
  //organization={Springer}
}

@article{zhou2019unet++,
  title={Unet++: Redesigning skip connections to exploit multiscale features in image segmentation},
  author={Zhou, Zongwei and Siddiquee, Md Mahfuzur Rahman and Tajbakhsh, Nima and Liang, Jianming},
  journal=TMI,
  volume={39},
  number={6},
  pages={1856--1867},
  year={2019},
  //publisher={ieee}
}

@inproceedings{fang2019selective,
  title={Selective feature aggregation network with area-boundary constraints for polyp segmentation},
  author={Fang, Yuqi and Chen, Cheng and Yuan, Yixuan and Tong, Kai-yu},
  booktitle=MICCAI,
  //pages={302--310},
  year={2019},
  //organization={Springer}
}

@inproceedings{zhang2020adaptive,
  title={Adaptive context selection for polyp segmentation},
  author={Zhang, Ruifei and Li, Guanbin and Li, Zhen and Cui, Shuguang and Qian, Dahong and Yu, Yizhou},
  booktitle=MICCAI,
  //pages={253--262},
  year={2020},
  //organization={Springer}
}

@inproceedings{patel2021enhanced,
  title={Enhanced u-net: A feature enhancement network for polyp segmentation},
  author={Patel, Krushi and Bur, Andr{\'e}s M and Wang, Guanghui},
  booktitle=CRV,
  //pages={181--188},
  year={2021},
  //organization={IEEE}
}

@inproceedings{wei2021shallow,
  title={Shallow attention network for polyp segmentation},
  author={Wei, Jun and Hu, Yiwen and Zhang, Ruimao and Li, Zhen and Zhou, S Kevin and Cui, Shuguang},
  booktitle=MICCAI,
  //pages={699--708},
  year={2021},
  //organization={Springer}
}

@inproceedings{kim2021uacanet,
  title={Uacanet: Uncertainty augmented context attention for polyp segmentation},
  author={Kim, Taehun and Lee, Hyemin and Kim, Daijin},
  booktitle=ACMMM,
  //pages={2167--2175},
  year={2021}
}

@article{hu2026pranet,
  title={Pranet-v2: Dual-supervised reverse attention for medical image segmentation},
  author={Hu, Bo-Cheng and Ji, Ge-Peng and Shao, Dian and Fan, Deng-Ping},
  journal=CVMJ,
  year={2026},
  volume={12},
  number={2},
  pages={493-500},
  //publisher={TUP}
}

@article{dong2023polyp,
  title={Polyp-PVT: Polyp Segmentation with Pyramid Vision Transformers},
  author={Dong, Bo and Wang, Wenhai and Fan, Deng-Ping and Li, Jinpeng and Fu, Huazhu and Shao, Ling},
  journal=JAIR,
  volume={2},
  pages={9150015},
  year={2023},
  //publisher={清华大学出版社}
}

@article{zhou2023cross,
  title={Cross-level Feature Aggregation Network for Polyp Segmentation},
  author={Zhou, Tao and Zhou, Yi and He, Kelei and Gong, Chen and Yang, Jian and Fu, Huazhu and Shen, Dinggang},
  journal=PR,
  volume={140},
  pages={109555},
  year={2023},
  publisher={Elsevier}
}

@inproceedings{yang2019my,
  title={Where is my mirror?},
  author={Yang, Xin and Mei, Haiyang and Xu, Ke and Wei, Xiaopeng and Yin, Baocai and Lau, Rynson WH},
  booktitle=ICCV,
  //pages={8809--8818},
  year={2019}
}

@article{bernal2015wm,
  title={WM-DOVA maps for accurate polyp highlighting in colonoscopy: Validation vs. saliency maps from physicians},
  author={Bernal, Jorge and S{\'a}nchez, F Javier and Fern{\'a}ndez-Esparrach, Gloria and Gil, Debora and Rodr{\'\i}guez, Cristina and Vilari{\~n}o, Fernando},
  journal=CMIG,
  volume={43},
  pages={99--111},
  year={2015},
  publisher={Elsevier}
}

@article{vazquez2017benchmark,
  title={A benchmark for endoluminal scene segmentation of colonoscopy images},
  author={V{\'a}zquez, David and Bernal, Jorge and S{\'a}nchez, F Javier and Fern{\'a}ndez-Esparrach, Gloria and L{\'o}pez, Antonio M and Romero, Adriana and Drozdzal, Michal and Courville, Aaron},
  journal=JHE,
  volume={2017},
  number={1},
  pages={4037190},
  year={2017},
  publisher={Wiley Online Library}
}

@inproceedings{jha2019kvasir,
  title={Kvasir-seg: A segmented polyp dataset},
  author={Jha, Debesh and Smedsrud, Pia H and Riegler, Michael A and Halvorsen, P{\aa}l and De Lange, Thomas and Johansen, Dag and Johansen, H{\aa}vard D},
  booktitle=MMM,
  //pages={451--462},
  year={2019},
  //organization={Springer}
}

@article{silva2014toward,
  title={Toward embedded detection of polyps in wce images for early diagnosis of colorectal cancer},
  author={Silva, Juan and Histace, Aymeric and Romain, Olivier and Dray, Xavier and Granado, Bertrand},
  journal=CARS,
  volume={9},
  number={2},
  pages={283--293},
  year={2014},
  publisher={Springer}
}

@article{tajbakhsh2015automated,
  title={Automated polyp detection in colonoscopy videos using shape and context information},
  author={Tajbakhsh, Nima and Gurudu, Suryakanth R and Liang, Jianming},
  journal=TMI,
  volume={35},
  number={2},
  pages={630--644},
  year={2015},
  //publisher={IEEE}
}

@inproceedings{han2026beyond,
  title={Beyond Appearance: Camouflaged Object Detection via Geometric Structure},
  author={Han, Jinyu and Wu, Changguang and Sun, Fuming and Tang, Jinhui},
  booktitle=CVPR,
  //pages={25830--25840},
  year={2026}
}

@article{sun2025unet,
  title={A UNet-Like Transformer Network for Camouflaged Object Detection},
  author={Sun, Fuming and Han, Jinyu and Wu, Weiyi and Sun, Jing and Wang, Mengyin and Li, Haojie},
  journal=TMM,
  volume={27},
  pages={9267--9280},
  year={2025}
}

@inproceedings{luo2024vscode,
  title={Vscode: General visual salient and camouflaged object detection with 2d prompt learning},
  author={Luo, Ziyang and Liu, Nian and Zhao, Wangbo and Yang, Xuguang and Zhang, Dingwen and Fan, Deng-Ping and Khan, Fahad and Han, Junwei},
  booktitle=CVPR,
  //pages={17169--17180},
  year={2024}
}

@article{luo2025vscode,
  title={VSCode-V2: Dynamic Prompt Learning for General Visual Salient and Camouflaged Object Detection With Two-Stage Optimization},
  author={Luo, Ziyang and Liu, Nian and Yang, Xuguang and Zhang, Dingwen and Fan, Deng-Ping and Khan, Fahad Shahbaz and Han, Junwei},
  journal=TPAMI,
  year={2025},
  volume={48},
  number={3},
  pages={3137-3153},
  //publisher={IEEE}
}

@inproceedings{sun2024frequency,
  title={Frequency-spatial entanglement learning for camouflaged object detection},
  author={Sun, Yanguang and Xu, Chunyan and Yang, Jian and Xuan, Hanyu and Luo, Lei},
  booktitle=ECCV,
  //pages={343--360},
  year={2024},
  //organization={Springer}
}

@inproceedings{das2025camouflage,
  title={Camouflage Anything: Learning to Hide using Controlled Out-painting and Representation Engineering},
  author={Das, Biplab and Gopalakrishnan, Viswanath},
  booktitle=CVPR,
  //pages={3603--3613},
  year={2025}
}

@article{le2019anabranch,
  title={Anabranch network for camouflaged object segmentation},
  author={Le, Trung-Nghia and Nguyen, Tam V and Nie, Zhongliang and Tran, Minh-Triet and Sugimoto, Akihiro},
  journal=CVIU,
  volume={184},
  pages={45--56},
  year={2019},
  publisher={Elsevier}
}

@inproceedings{lin2020progressive,
  title={Progressive mirror detection},
  author={Lin, Jiaying and Wang, Guodong and Lau, Rynson WH},
  booktitle=CVPR,
  //pages={3697--3705},
  year={2020}
}

@inproceedings{he2023efficient,
  title={Efficient mirror detection via multi-level heterogeneous learning},
  author={He, Ruozhen and Lin, Jiaying and Lau, Rynson WH},
  booktitle=AAAI,
  //volume={37},
  //number={1},
  //pages={790--798},
  year={2023}
}

@inproceedings{guan2022learning,
  title={Learning semantic associations for mirror detection},
  author={Guan, Huankang and Lin, Jiaying and Lau, Rynson WH},
  booktitle=CVPR,
  //pages={5941--5950},
  year={2022}
}

@inproceedings{li2024sizeinvariance,
  title={Size-invariance Matters: Rethinking Metrics and Losses for Imbalanced Multi-object Salient Object Detection}, 
  author={Feiran Li and Qianqian Xu and Shilong Bao and Zhiyong Yang and Runmin Cong and Xiaochun Cao and Qingming Huang},
  booktitle=ICML,
  //pages={},
  year={2024}
}

@article{mokrzycki2011colour,
  title={Colour difference {$\Delta E$} a survey},
  author={Mokrzycki, WS and Tatol, Maciej},
  journal={Mach. Graph. Vis},
  volume={20},
  number={4},
  pages={383--411},
  year={2011}
}

@article{stevens2009animal,
  title={Animal camouflage: current issues and new perspectives},
  author={Stevens, Martin and Merilaita, Sami},
  journal={Philosophical Transactions of the Royal Society B: Biological Sciences},
  pages={423--427},
  year={2009},
  publisher={The Royal Society},
  volume={364},
  number={1516}, 
}

@inproceedings{zhai2021mutual,
  title={Mutual graph learning for camouflaged object detection},
  author={Zhai, Qiang and Li, Xin and Yang, Fan and Chen, Chenglizhao and Cheng, Hong and Fan, Deng-Ping},
  booktitle=CVPR,
  //pages={12997--13007},
  year={2021}
}

@inproceedings{zhong2022detecting,
  title={Detecting camouflaged object in frequency domain},
  author={Zhong, Yijie and Li, Bo and Tang, Lv and Kuang, Senyun and Wu, Shuang and Ding, Shouhong},
  booktitle=CVPR,
  //pages={4504--4513},
  year={2022}
}

@inproceedings{jia2022segment,
  title={Segment, magnify and reiterate: Detecting camouflaged objects the hard way},
  author={Jia, Qi and Yao, Shuilian and Liu, Yu and Fan, Xin and Liu, Risheng and Luo, Zhongxuan},
  booktitle=CVPR,
  //pages={4713--4722},
  year={2022}
}

@inproceedings{liu2022boosting,
  title={Boosting camouflaged object detection with dual-task interactive transformer},
  author={Liu, Zhengyi and Zhang, Zhili and Tan, Yacheng and Wu, Wei},
  booktitle=ICPR,
  //pages={140--146},
  year={2022},
  organization={IEEE}
}

@article{zhang2023tprnet,
  title={Tprnet: camouflaged object detection via transformer-induced progressive refinement network},
  author={Zhang, Qiao and Ge, Yanliang and Zhang, Cong and Bi, Hongbo},
  journal={The Visual Computer},
  volume={39},
  number={10},
  pages={4593--4607},
  year={2023},
  publisher={Springer}
}

@inproceedings{jafari2016skin,
  title={Skin lesion segmentation in clinical images using deep learning},
  author={Jafari, Mohammad H and Karimi, Nader and Nasr-Esfahani, Ebrahim and Samavi, Shadrokh and Soroushmehr, S Mohamad R and Ward, K and Najarian, Kayvan},
  booktitle=ICPR,
  //pages={337--342},
  year={2016},
  //organization={IEEE}
}

@article{wang2026expose,
  title={Expose camouflage in the water: underwater camouflaged instance segmentation and dataset},
  author={Wang, Chuhong and Li, Hua and Li, Chongyi and Liu, Huazhong and Tang, Xiongxin and Kwong, Sam},
  journal=TIP,
  volume={35},
  pages={3283--3298},
  year={2026},
  publisher={IEEE}
}

@inproceedings{kalra2020deep,
  title={Deep polarization cues for transparent object segmentation},
  author={Kalra, Agastya and Taamazyan, Vage and Rao, Supreeth Krishna and Venkataraman, Kartik and Raskar, Ramesh and Kadambi, Achuta},
  booktitle=CVPR,
  //pages={8602--8611},
  year={2020}
}

@inproceedings{perazzi2012saliency,
  title={Saliency filters: Contrast based filtering for salient region detection},
  author={Perazzi, Federico and Kr{\"a}henb{\"u}hl, Philipp and Pritch, Yael and Hornung, Alexander},
  booktitle=CVPR,
  //pages={733--740},
  year={2012},
  organization={IEEE}
}

@inproceedings{achanta2009frequency,
  title={Frequency-tuned salient region detection},
  author={Achanta, Radhakrishna and Hemami, Sheila and Estrada, Francisco and Susstrunk, Sabine},
  booktitle=CVPR,
  //pages={1597--1604},
  year={2009},
  organization={IEEE}
}

@article{haralick1987image,
  title={Image analysis using mathematical morphology},
  author={Haralick, Robert M and Sternberg, Stanley R and Zhuang, Xinhua},
  journal=TPAMI,
  number={4},
  pages={532--550},
  year={1987},
  publisher={IEEE}
}

@article{luo2001development,
  title={The development of the CIE 2000 colour-difference formula: CIEDE2000},
  author={Luo, M Ronnier and Cui, Guihua and Rigg, Bryan},
  journal={Color Research \& Application},
  volume={26},
  number={5},
  pages={340--350},
  year={2001},
  publisher={Wiley Online Library}
}
}

\newpage

\end{document}